%% file: tmlr.tex
\documentclass[10pt]{article}
\usepackage[accepted]{tmlr}

\input{math_commands.tex}

\usepackage{hyperref}
\usepackage{url}
\usepackage{graphicx}
\usepackage{booktabs}
\usepackage{tabularx}
\usepackage{xeCJK}

\title{Steering the Language Axis: From Linear Decodability to Causal Control}

\author{\name Arnav Srivastav \email asriva31@ucsc.edu \\
      \addr University of California, Santa Cruz}
\def\month{11}
\def\year{2026}

\providecommand{\openreview}{}
\begin{document}

\maketitle

\begin{abstract}

Despite the impressive multilingual capabilities of Large Language Models, the latent dynamics dictating
language selection remain poorly understood. In this work, we ask whether language identity is merely
linearly decodable from hidden states, or if it can be causally controlled by a compact activation direction.
We conduct a large-scale causal intervention analysis across seven model instances spanning roughly 1B--27B parameters and four model families, comprising approximately 1.32 million generated sequences. We isolate PCA-derived ``language axes'' to perform steering and ablation experiments across English--Chinese, English--Spanish, and Chinese--Spanish settings.

Steering along these geometric directions reliably alters output language across models and language pairs, demonstrating that linearly decodable language structure can also be causally effective during generation. However, this control is not governed by a single universal layer or fixed steering pattern. Instead, our layerwise analysis reveals that effective intervention regions vary systematically with model family, language pair, steering direction, and scale.

Our results show that steering toward English is more stable than steering away from it across English-centered settings, while the Chinese--Spanish experiments show that this asymmetry does not generalize to all language pairs. Furthermore, we show that aggressive late-layer interventions eventually cause the model to enter a degeneration regime, though the onset and nature of this failure depend on the specific model and direction.
Through BERTScore, LLM-as-a-judge evaluation, and ARC-Easy task-based evaluation, we show that language control can preserve meaning and performance, though a balance exists: aggressive or poorly placed steering can trigger semantic drift and failures on tasks. Ultimately, these findings demonstrate that language identity is not only linearly decodable, but causally effective as a structured, direction-dependent, and layer-specific feature of multilingual model representations.

\end{abstract}

\section{Introduction}
Since the introduction of the foundational Transformer architecture \citep{vaswani2017}, large language models have evolved from narrow, task-specific systems into highly capable general-purpose reasoning 
engines. Driven by massive scale and diverse pretraining corpora, modern decoder-only transformers now 
display impressive multilingual capabilities \citep{brown2020,workshop2023bloom176bparameteropenaccessmultilingual}. Yet the internal
mechanisms that determine which language a model chooses to generate remain poorly 
understood. When a multilingual model responds in English rather than Chinese, or Spanish rather than
English, is this decision controlled by a diffuse network-wide pattern, a late-stage decoding bias, or a 
compact, steerable latent variable?

Answering this question is critical for both mechanistic interpretability and practical reliability.
Scientifically, language selection provides a natural testbed for studying how transformer models
separate semantic content from surface form. A model processing two sentences with similar meaning but in 
different languages may represent both the shared semantic structure and the language-specific syntax.
Identifying exactly where and how these components are represented clarifies whether multilingual 
models rely on language-neutral semantic hubs, language-anchored manifolds, or direction-specific transformations between linguistic subspaces. Practically, output language control is a frequent failure
mode in deployed systems, as models often default to English, unexpectedly code switch, or ignore
explicit instructions to answer in a target language \citep{marchisio-etal-2024-understanding,nie-etal-2025-mechanistic,wang-etal-2025-language-mixing,oh2026olaoutputlanguagealignment}. A mechanistic account of language control could enable more reliable multilingual generation, without the need for expensive fine-tuning.

Prior work has demonstrated that language identity is often linearly decodable from hidden states \citep{kim2025languagedirectionsaligntoken,srinivasan2023counterfactuallyprobinglanguageidentity}.
Linear probes, representation clustering, and cross-lingual embedding analyses all
suggest that transformer activations contain rich language-specific structure \citep{mousi-etal-2024-exploring,chang-etal-2022-geometry}. However,
decodability alone does not equal causal use. A feature might be easy to recover from activations, while playing little to no role in actual generation. Conversely, a specific activation direction might correlate
with language identity but fail to alter the model's behavior when intervened upon. Distinguishing correlation from true causal control requires direct activation intervention by modifying the model's 
internal state during inference---an approach recently shown to successfully mitigate language confusion at the neuron level \citep{nie-etal-2025-mechanistic}, and measuring whether the output language reliably changes.

In this work, we conduct a causal intervention analysis across seven model instances spanning roughly 1B--27B parameters and four model families, to determine if language identity functions as a manipulable
latent variable. By extracting pair-specific, PCA-derived "language axes" from held-out activations, we directly steer and ablate the model's residual-stream states during inference. To ensure our findings generalize beyond a single architecture, scale, script, or language pairing, we evaluate interventions across English--Chinese, English--Spanish, and Chinese--Spanish settings. By testing a wide range of layers and steering strengths, we can determine if these language representations are causally effective, but also how their efficacy varies across model family, scale, language pair, and steering direction. Ultimately, this approach allows us to move beyond observational probing, providing direct evidence that multilingual generation is governed by causally active, direction-dependent, and layer-specific geometry.

\subsection{Contributions}
Our core contributions are as follows:
\begin{itemize}

\item \textbf{Causal Methodology and Large-Scale Identification:} By isolating PCA-derived ``language axes'' across seven model variants spanning roughly 1B--27B parameters and four model families, we demonstrate that these directions generalize across different architectures and can successfully steer model generation. We further validate that these effects are specific to language steering rather than generic model disruption through a series of control experiments.

\item \textbf{Mechanistic Insights and Cross-Lingual Behavior:} Through large-scale, full-depth intervention sweeps across English--Chinese, English--Spanish, and Chinese--Spanish, we show that language-control geometry is explicitly model-, language-pair-, direction-, and scale-dependent. Instead of finding a single universal layer, we find that causal sensitivity is concentrated in specific, structured regions that shift depending on the setting.

\item \textbf{Directional Differences in Language Steering:} Across English-centered settings, we find that steering toward English is consistently more stable than steering away from it. However, experiments on Chinese--Spanish show that this pattern does not generalize to all language pairs, suggesting that steering behavior depends strongly on the languages involved.

\item \textbf{Preserving Meaning and Task Performance:} We evaluate whether changing a model's output language also changes what the model is trying to say or how well it performs. Using BERTScore, LLM-as-a-judge evaluation, ARC-Easy task evaluation, and late-layer diagnostics, we find that language steering can preserve semantic content and task behavior within suitable intervention regions. However, stronger or poorly placed interventions can instead produce semantic drift, output-format failures, repetitive generation, or model degeneration.

\end{itemize}

\section{Related Work}
\subsection{Multilingual Representations and Linear Probing}
Large language models acquire substantial multilingual capabilities through large-scale pretraining on
heterogeneous text corpora \citep{brown2020,workshop2023bloom176bparameteropenaccessmultilingual}. Prior work
has shown that multilingual models often organize semantically similar sentences across languages into partially shared representational spaces, while also preserving language-specific structure such as script, 
morphology, syntax and tokenization artifacts \citep{chang-etal-2022-geometry,zhang2024differentstructuralsimilaritiesdifferences, verma2026multilinguallanguagemodelsencode}.
A growing body of interpretability work studies these representations through the lens of linear features,
showing that complex attributes---such as factual knowledge, refusal behavior, and language identity---are often
linearly decodable from intermediate activations \citep{kim2025languagedirectionsaligntoken,srinivasan2023counterfactuallyprobinglanguageidentity,marks2024geometrytruthemergentlinear,arditi2024refusallanguagemodelsmediated,zou2025representationengineeringtopdownapproach}. However, the interpretation of linear probes is subtle. A successful probe only demonstrates that 
information is present in the representation; it does not establish that the model uses that information
causally during generation \citep{ravichander-etal-2021-probing}. A feature might be easy to decode from hidden states, while
playing no active role in the model's output. Our work bridges this gap by moving from representational
analysis to direct causal intervention.

\subsection{Activation Steering and Causal Interventions}
Activation steering has recently emerged as a method for controlling model behavior without altering 
model weights. By adding a vector corresponding to a target attribute to hidden states during inference, 
researchers can predictably manipulate
generation \citep{li2024inferencetimeinterventionelicitingtruthful,turner2024steeringlanguagemodelsactivation,lu2026assistantaxissituatingstabilizing}. Related techniques---such as activation
addition, representation engineering, and targeted ablation---provide a framework to test whether an internal
representation is causally sufficient for a specific behavior \citep{turner2024steeringlanguagemodelsactivation,meng2023locatingeditingfactualassociations,zou2025representationengineeringtopdownapproach}. Contrastive Activation Addition (CAA) similarly constructs low-dimensional steering directions from contrastive activations \citep{rimsky-etal-2024-steering}. We compare CAA with our PCA-based language directions under matched perturbation magnitude as an additional steering baseline.

\subsection{Language Control and Multilingual Reliability}
Understanding the internal mechanisms of language selection is essential for clarifying why multilingual models 
exhibit specific behavioral patterns. It is well documented that models frequently default to high-resource languages
like English, unexpectedly code switch, or struggle to adhere to the target language \citep{marchisio-etal-2024-understanding,oh2026olaoutputlanguagealignment,nie-etal-2025-mechanistic,wang-etal-2025-language-mixing}.
Current solutions rely on prompt engineering, multilingual fine-tuning, or preference tuning \citep{marchisio-etal-2024-understanding}. Other work has also used language directions for inference-time mitigation of unintended code-switching \citep{goncharov2026languagesteeringlatentspace}.

We go beyond simply asking if language-specific directions can influence a model's output; instead, we investigate the causal geometry of language identity across a wide range of models, scales, and language pairs. By combining full-depth intervention sweeps with semantic and task-level analyses, we characterize not just where these interventions succeed, but also how their effects change across settings and when they begin to cause model disruption. Ultimately, this work provides a more complete mechanistic account of how multilingual transformers represent and control language identity.


\section{Methodology and Experimental Setup}
\subsection{Models and Datasets}

We conduct our causal intervention analysis across seven decoder-only language model
instances spanning roughly 1B--27B parameters and four model families. Our experiments
include Qwen 3.5-2B (24 layers, $d = 2048$),
Qwen 3.5-9B (32 layers, $d = 4096$) ,
Qwen 3.5-27B (64 layers, $d = 5120$),
Llama-3.2-1B-Instruct (16 layers, $d = 2048$),
Llama-3.1-8B (32 layers, $d = 4096$),
Gemma-4-12B-it (48 layers, $d = 3840$), and
Granite-4.1-8B  (40 layers, $d = 4096$) \citep{qwen35blog,grattafiori2024llama3herdmodels,ibm_granite_huggingface,gemma2026gemma4}.
All interventions modify hidden states during the forward pass at inference time;
model weights remain completely frozen.

We evaluate generation using the FLORES-200 dataset \citep{nllbteam2022languageleftbehindscaling}. Our primary experiments study bidirectional English--Chinese (EN$\leftrightarrow$ZH) language control. We additionally evaluate English--Spanish (EN$\leftrightarrow$ES) as a same-script setting and Chinese--Spanish (ZH$\leftrightarrow$ES) as a non-English-centered setting. To calibrate the language axes, we extract hidden states from held-out parallel sentence pairs using the MultiUN corpus \citep{eisele-chen-2010-multiun}. For the smaller-model experiments, we use 20,000 valid calibration sentence pairs per language pair, while experiments on the larger models use 10,000 pairs. Calibration and evaluation are performed on separate corpora.

For Qwen-9B, Llama-8B, Gemma-12B, and Granite-8B, we perform full-depth intervention sweeps in both EN$\rightarrow$ZH and ZH$\rightarrow$EN directions. Due to its substantially greater computational cost, the Qwen-27B experiment instead evaluates a targeted subset of layers and is treated as evidence of scale generalization rather than a full-depth replication.

\subsection{Isolating the Language Axis}
For each layer $\ell$, we collect hidden states $h_{\ell,i}$ from the calibration sentences in both the source and target languages. Each calibration sentence is represented using the hidden state at its final non-padding token. We center the activations by subtracting the layerwise calibration mean $\mu_\ell$:

$$
\tilde{h}_{\ell,i} = h_{\ell,i} - \mu_\ell
$$

We then fit Principal Component Analysis (PCA) to the centered activations, retaining the top 150 components. To identify the specific component most associated with language separation, we calculate the Cohen's $d$ effect size between the source and target language distributions along each principal component. We select the component with the maximum absolute Cohen's $d$ and orient it so that a positive projection corresponds to the target language. We denote this resulting unit vector as $d_\ell$. The scalar projection of a hidden state onto this axis provides a one-dimensional language coordinate for that specific layer:

$$
z_\ell(h) = (h - \mu_\ell)^\top d_\ell
$$
\subsection{Causal Interventions}

To test the causal efficacy of the language axis, we perform targeted interventions on the residual stream at individual layers.

\textbf{Steering:} We steer the model by adding a scaled displacement along the language axis. At layer $\ell$, the intervened hidden state is defined as:

$$
h'_{\ell} = h_{\ell} + \alpha \cdot \Delta_{\ell, s \to t} \cdot d_{\ell}
$$

where $\alpha \in [0, 30.0]$ is the steering strength multiplier and $\Delta_{\ell, s \to t}$ is the mean source-to-target scalar displacement estimated from the calibration data.

\textbf{Ablation:} To test the model's dependence on the identified language axis, we ablate it by projecting out the language component. This preserves all orthogonal components while mathematically zeroing out the targeted language coordinate:

$$
h'_{\ell} = h_{\ell} - ((h_{\ell} - \mu_{\ell})^\top d_{\ell})d_{\ell}
$$

\textbf{Matched Random Controls:} To ensure observed effects are direction-specific rather than caused by generic off-manifold noise, we apply random perturbations as a falsification test. We sample random unit vectors $r_\ell$ uniformly from the unit sphere. Due to the concentration of measure in high-dimensional spaces, these random
vectors are nearly orthogonal to the language axis by default:

$$
r_\ell^\top d_\ell \approx 0
$$

We then scale these random vectors to match the exact mathematical norm of the targeted steering intervention:

$$
h'_{\ell,rand} = h_\ell + \|\Delta h_{\ell,\alpha}\| r_\ell
$$

\textbf{Experimental Controls and Metrics}

Across approximately 1.32 million total generated sequences, we test intervention robustness against multiple prompt templates (default chat, English system prompt, target language system prompt, and raw completion) and intervention modes (prefill-only, decode-only, and always-on). Prefill-only steering is applied to the final prompt token, decode-only steering is applied during autoregressive decoding, and always-on steering uses both. Unless otherwise stated, our main sweeps use always-on steering. Generation is deterministic with \texttt{do\_sample=False} and \texttt{max\_new\_tokens=40}, with default beam settings and no top-$p$, top-$k$, or repetition penalties. For each prompt and steering direction, we generate one unsteered baseline using the same decoding settings and reuse it across intervention configurations. We evaluate the generated outputs using the following metrics:

\begin{itemize}
\item \textbf{Target-Language Ratio ($R_{target}$):} For experiments involving
Chinese, we measure the fraction of generated characters belonging to the
Chinese script, oriented according to the steering direction. For
English--Spanish, we measure the target-language probability assigned by the
\texttt{langid} classifier \citep{lui-baldwin-2012-langid}.
    
    \item \textbf{Repetition ($Rep$) and Entropy ($H$):} To detect model degeneration and evaluate generation diversity, we measure the output repetition rate via repeated n-grams, and calculate the token-level Shannon entropy.
    
    \item \textbf{Perturbation Magnitude ($\rho$):} To quantify the intervention dosage and account for the size of the imposed activation edit relative to the original state, we compute:
\end{itemize}

$$
\rho_{\ell,\alpha} = \frac{\|h'_{\ell} - h_{\ell}\|}{\|h_{\ell}\|}
$$

\textbf{Semantic Preservation:} To measure how strongly steering changes the semantic content of the model's baseline response, we compute BERTScore \citep{zhang2020bertscore} between the unsteered and steered generations. The unsteered output serves as a model-relative reference rather than an external ground-truth response.  We then supplement this with a blinded, multilingual "LLM-as-a-judge" evaluation on 800 response pairs. This judge evaluates semantic preservation, factual consistency, intent, fluency, and signs of degenerate behavior. The two responses are presented in random order, and the evaluator is kept unaware of the steering condition, model, intervention layer and strength, or our hypothesis. We additionally repeat 83 examples with the response ordering reversed as a consistency check. 

\textbf{Task-Level Evaluation:} We further test whether steering preserves task performance using 200 ARC-Easy questions. We compare unsteered generation, target-language prompting, and EN$\rightarrow$ZH steering at moderate and strong intervention strengths. To get a clear picture, we report two types of accuracy: end-to-end accuracy and accuracy conditioned on valid formatting. This distinction helps us see if a failure is a true task error or simply a case of the steering causing the model to lose its ability to format a proper answer.

We also compare our PCA directions against CAA \citep{rimsky-etal-2024-steering}, matching interventions by perturbation magnitude rather than raw steering strength.

\section{Results}
\subsection{Overview}

We evaluate whether PCA-derived language axes causally control output language across seven model instances spanning 1B--27B parameters and four model families, and across English--Chinese, English--Spanish, and Chinese--Spanish language pairs. 

\begin{enumerate}

\item \textbf{Language axis causality generalizes:} Adding the language-axis direction reliably induces target-language output across Qwen, Llama, Gemma, and Granite models, with the effect persisting from roughly 1B to 27B parameters.

\item \textbf{Layerwise control is structured but not universal:} Effective control regions vary across architecture, scale, language pair, and steering direction. Rather than identifying a single universal control layer, we find model-specific regions of causal sensitivity.

\item \textbf{Steering exhibits directional asymmetry:} Across the tested English-centered settings, steering toward English is consistently more stable than steering away from it. This pattern does not generalize consistently to Chinese--Spanish.

\item \textbf{Late layers remain fragile and sensitive:} Strong interventions in later layers frequently produce repetitive or otherwise degenerate generation, though the exact onset and failure pattern vary across models and steering directions.

\item \textbf{Successful steering can preserve model behavior:} Semantic and task-level evaluations show that language control can preserve meaning and task performance under appropriate intervention settings, while stronger or poorly placed steering can introduce semantic drift, formatting failures, and degeneration.

\end{enumerate}

\textit{Note on Supplementary Material:} Due to space constraints, we present representative layerwise results and cross-model summaries in the main text. Additional English--Spanish and Chinese--Spanish steering results are provided in the Appendix.

\subsection{Qwen English--Chinese: Bidirectional Steering and Layer Regimes}

We first examine bidirectional English--Chinese steering in Qwen 3.5-2B. Without steering ($\alpha=0$), the model produces almost no Chinese output. As steering strength increases in the $EN \rightarrow ZH$ direction, strong language control emerges primarily in the later layers. Layers 17--23 contain the clearest region of successful steering, where the model produces high levels of Chinese output while remaining relatively stable. We also observe a noticeable dip in steering effectiveness around Layer 12. At the final layer, very strong interventions can still force the model into Chinese, but these settings are much more prone to repetitive or otherwise degenerate generation.

\begin{figure}[htbp]
    \centering
    \includegraphics[width=\linewidth]{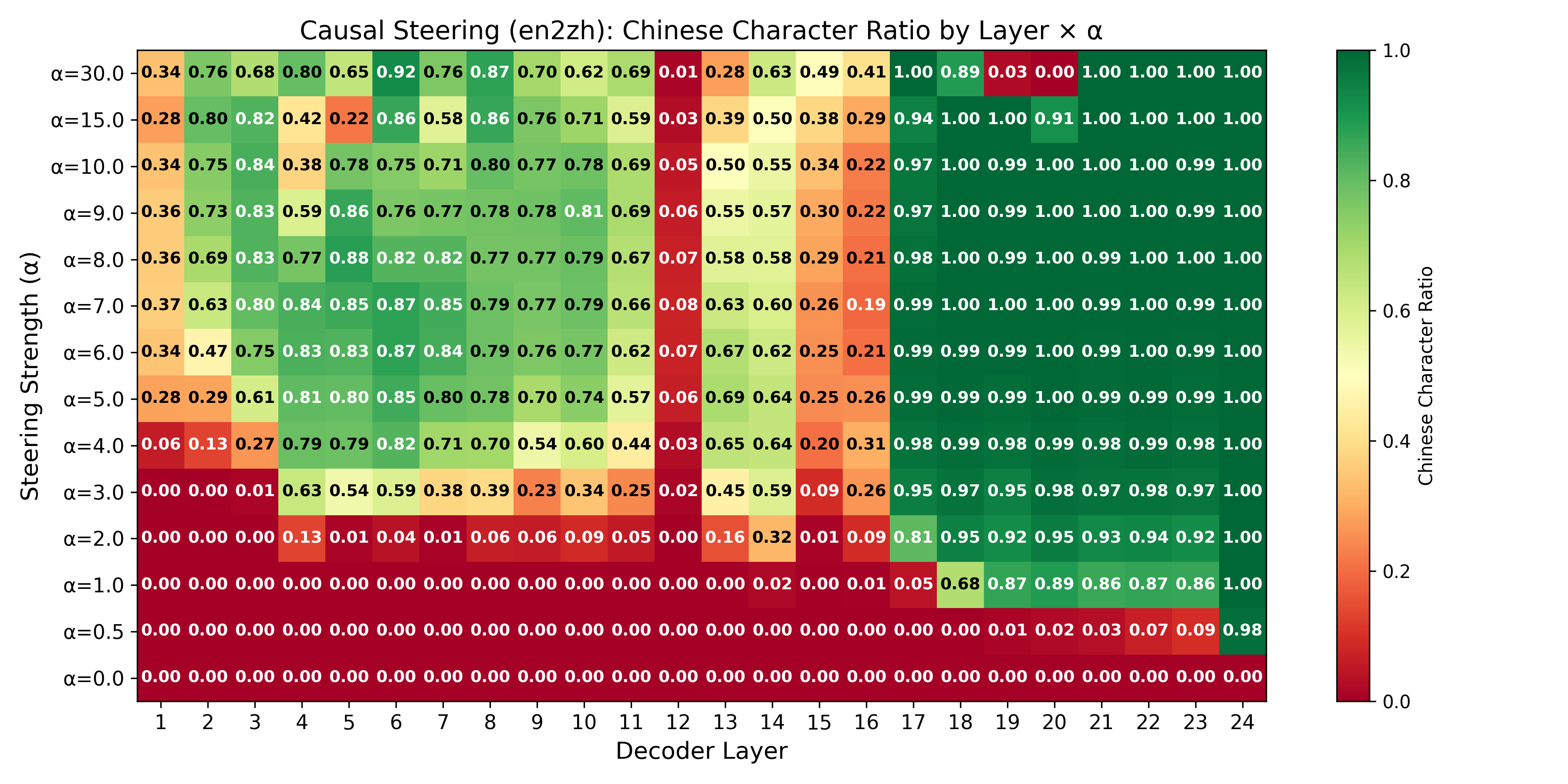} 
    \caption{\textbf{English-to-Chinese Steering.} Target-language ratio by layer and steering strength ($\alpha$) in Qwen 3.5-2B. Early layers are comparatively resistant to intervention, while strong language control emerges across later layers. A local region of reduced steering efficacy appears near Layer 12.}
    \label{fig:heatmap_en2zh_qwen}
\end{figure}

In the reverse $ZH \rightarrow EN$ direction, steering works across a much broader range of layers. The baseline Chinese-character ratio of 0.890 drops to just 0.003 at Layer 18 with $\alpha=2.0$, corresponding to almost entirely English output. Unlike $EN \rightarrow ZH$, this direction does not show the same mid-layer drop in steering effectiveness.

\begin{figure}[htbp]
    \centering
    \includegraphics[width=\linewidth]{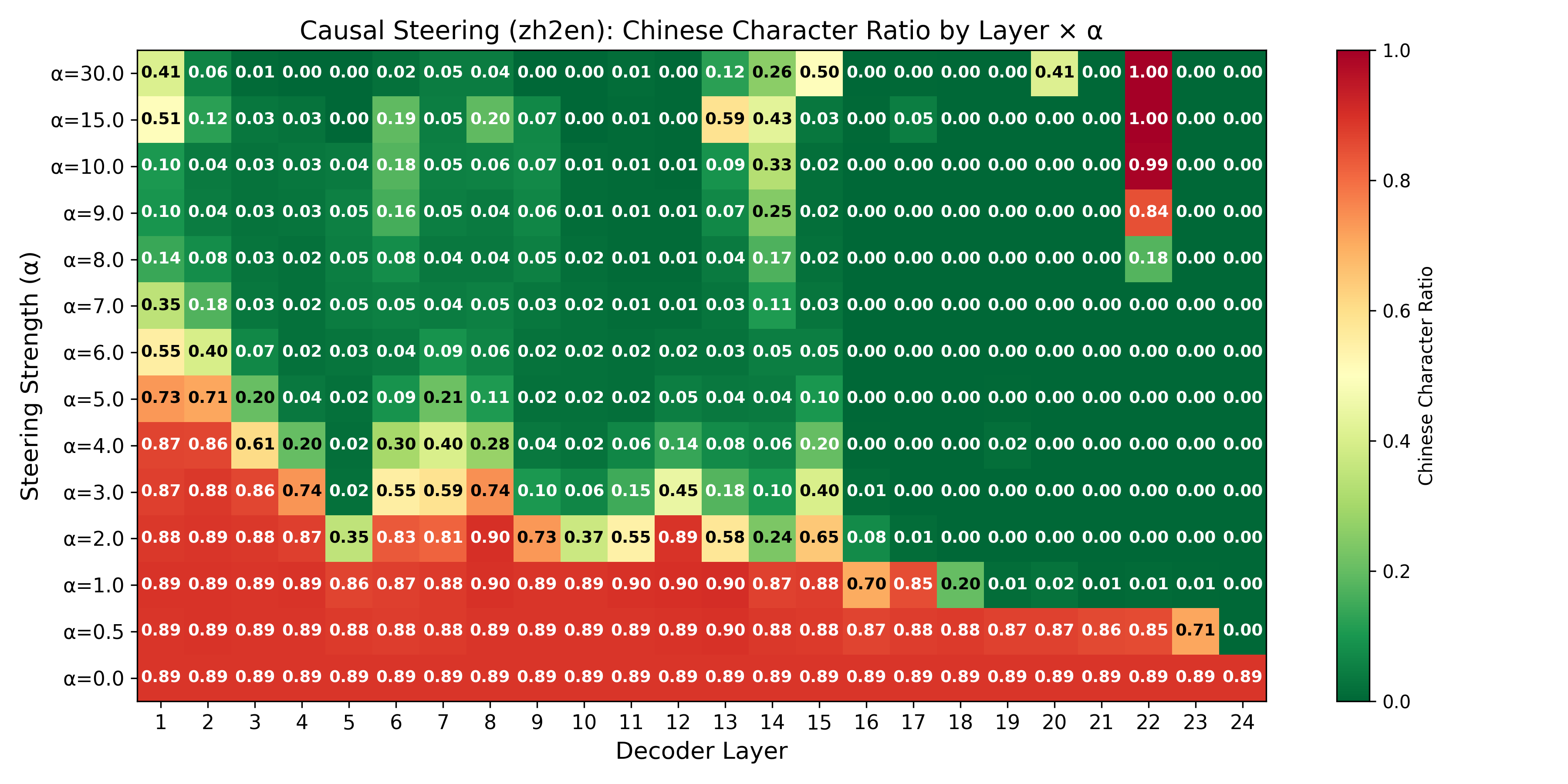} 
    \caption{\textbf{Chinese-to-English Steering.} Chinese-character ratio by layer and steering strength ($\alpha$) in Qwen 3.5-2B. Steering toward English succeeds across a broader portion of the network than the reverse $EN \rightarrow ZH$ direction.}
    \label{fig:heatmap_zh2en_qwen}
\end{figure}

Taken together, these results show a clear asymmetry between the two steering directions. In Qwen 3.5-2B, steering toward English works across a broader range of layers and strengths than steering away from English. As we show in Section~\ref{sec:directional_asymmetry}, the same pattern also appears in other model families and English-centered language pairs.

\subsection{Causal Specificity and Ablation}

Norm-matched random controls test whether the observed language switching comes from the language direction itself, rather than from simply perturbing the model’s hidden states. If this were the case, random directions of equal norm should produce changes comparable to those induced by the language axis. In Qwen $EN \rightarrow ZH$ at Layer 18 and $\alpha=5.0$, language-axis steering reaches a Chinese-character ratio of 0.994, whereas the matched random intervention reaches only 0.027. In the reverse direction, language-axis steering reduces the Chinese-character ratio to 0.001, compared with 0.866 under the matched random control. Thus, in these settings, the identified language direction produces substantially stronger language control than an equal-norm random perturbation. Additional random-control comparisons are provided in Appendix C.

\begin{figure}[htbp]
    \centering
    \includegraphics[width=0.8\textwidth]{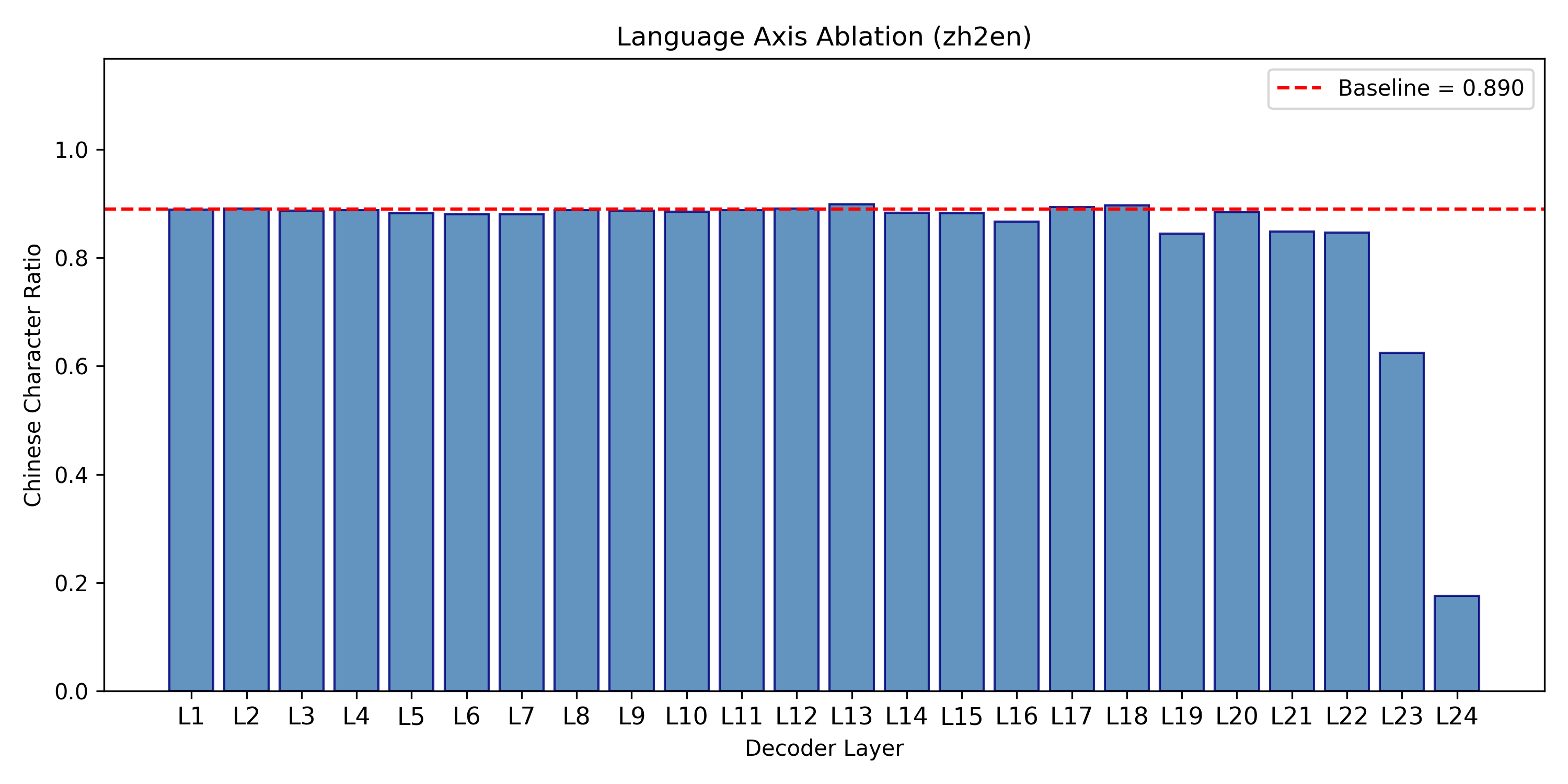}
    \caption{\textbf{Qwen 3.5-2B Language Axis Ablation ($ZH \rightarrow EN$).} Chinese-character ratio after ablating the identified language component at each decoder layer. The strongest shift toward English occurs at the final layer.}
    \label{fig:ablation_zh2en}
\end{figure}

Ablation provides complementary evidence that the identified direction contributes causally to language selection. In Qwen $ZH \rightarrow EN$, ablating the language-axis component in early and middle layers leaves generation close to the 0.890 Chinese baseline, whereas ablation at Layer 24 reduces the Chinese-character ratio to 0.176. Removing the late-layer component therefore produces a strong shift toward English in this setting. We interpret this as evidence of an English-directed fallback under ablation, rather than as proof that the single identified axis is strictly necessary for language identity.

\subsection{Same-Script Replication with English-Spanish and Cross-Script Chinese--Spanish}

We next test whether the same steering behavior extends beyond English--Chinese. English--Spanish gives us a same-script comparison, while Chinese--Spanish lets us study language control without English appearing in either direction.

For English--Spanish, steering works in both directions for Qwen 3.5-2B and Llama-3.2-1B. In Qwen, $EN \rightarrow ES$ reaches a Spanish target-language ratio of 1.000 at Layer 17 with $\alpha=3$, while several $ES \rightarrow EN$ settings produce almost entirely English output. The layerwise pattern, however, looks quite different from English--Chinese. In Qwen, $EN \rightarrow ES$ shows two clear regions of strong steering around Layers 5--7 and 17--21, while $ES \rightarrow EN$ is concentrated more heavily in the middle layers. Llama also supports bidirectional English--Spanish steering, with a broader region of effectiveness across the early and middle layers.

We then test Chinese--Spanish to see whether the effect still holds when English is removed entirely, and steering remains effective. In Qwen, $ES \rightarrow ZH$ reaches complete target-language control at Layer 20 with $\alpha=1$ while maintaining low repetition. We also find successful steering in the reverse $ZH \rightarrow ES$ direction, although the range of effective settings is narrower.

Taken together, these results show that PCA-derived language directions are not limited to English-centered or cross-script transitions. At the same time, the layers where steering works best continue to shift across language pairs and directions, suggesting that there is no single universal location for language control. The strong asymmetry we observe when steering toward English also does not appear in the same way for Chinese--Spanish.

Additional English--Spanish and Chinese--Spanish steering results are provided in the Appendix.

\subsection{Generalization Across Model Families and Scale}

Our small-model experiments show that causal language steering works in both Qwen and Llama, but it is still unclear whether the same effect holds at larger scales and across additional model families. To test this, we extend the English--Chinese experiments to Qwen 3.5-9B, Llama-3.1-8B, Gemma-4-12B, and Granite-4.1-8B, sweeping the full model depth in both steering directions. We also test a targeted set of layers in Qwen 3.5-27B.

The effect persists across all four larger models. Each model reaches complete target-language control in at least one layer--strength setting for both $EN \rightarrow ZH$ and $ZH \rightarrow EN$. Together with the Qwen 3.5-2B and Llama-3.2-1B results, this shows that causal language steering generalizes across four model families and across roughly an order of magnitude in model scale.

\begin{figure}[htbp]
    \centering
    \includegraphics[width=\linewidth]{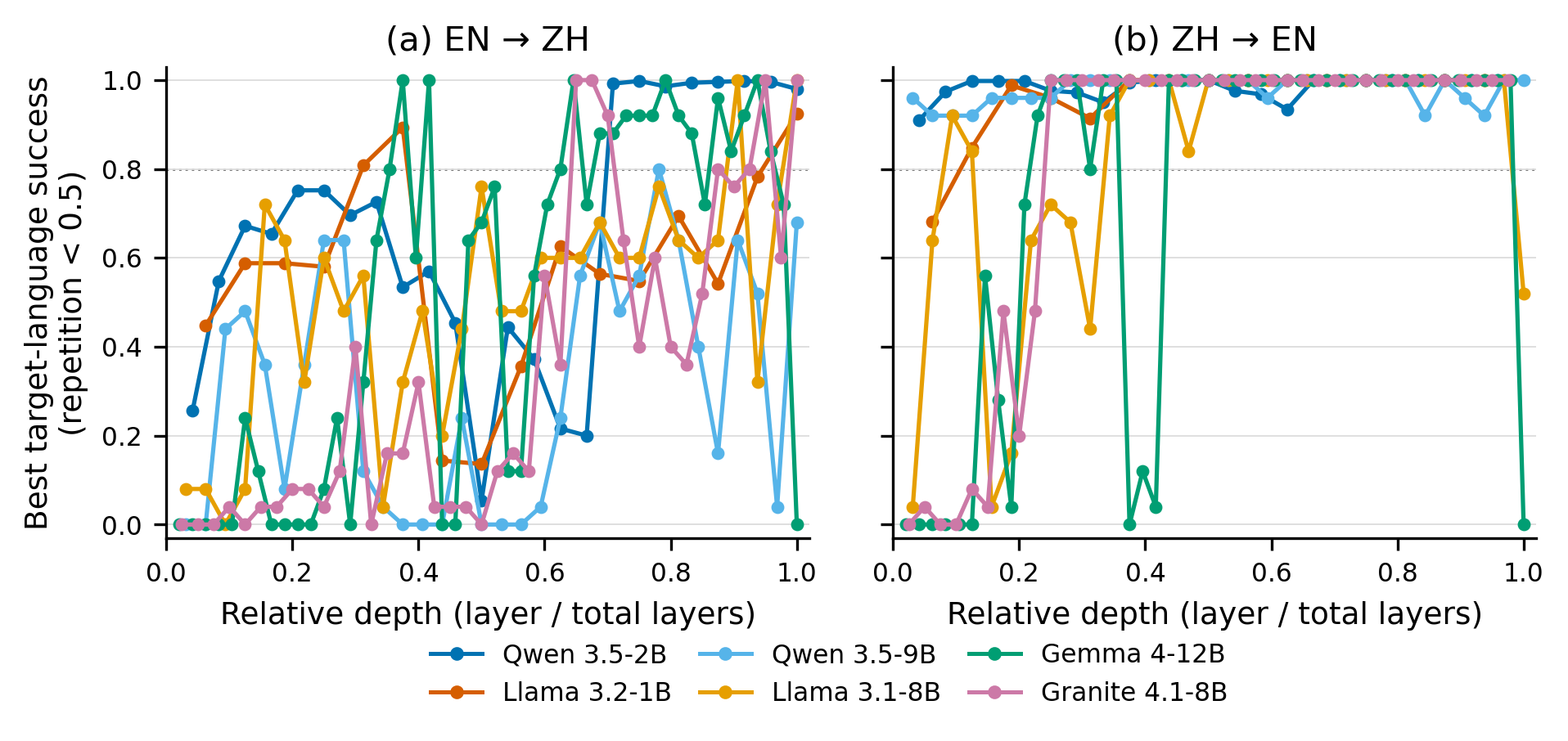}
    \caption{\textbf{Cross-model layerwise language control.}
    Best target-language performance among nondegenerate interventions across relative decoder depth for (a) $EN \rightarrow ZH$ and (b) $ZH \rightarrow EN$. Steering works across model families, but the layers where it is most effective differ substantially across models and directions.}
    \label{fig:cross_model_layerwise}
\end{figure}

Figure~\ref{fig:cross_model_layerwise} shows that although causal steering generalizes across models, the layerwise pattern is far from universal. For $EN \rightarrow ZH$, the strongest intervention regions appear at different depths depending on the model. Some models have relatively narrow windows where steering works well, while others remain steerable across a much broader portion of the network. In comparison, $ZH \rightarrow EN$ is generally easier to control, with strong target-language behavior extending across more of the decoder depth in most models.

The targeted Qwen 3.5-27B experiment suggests that this effect also persists at larger scale. We continue to observe strong $ZH \rightarrow EN$ steering in the layers we test, while $EN \rightarrow ZH$ is weaker in this limited sample. Because we do not sweep the full depth of the 27B model, we treat this result as additional evidence of scale generalization rather than as a direct layerwise comparison.

Overall, these results show that the causal effect itself is broadly consistent across architectures and scales, even though the layers where steering works best remain highly dependent on the model and steering direction.

\subsection{Directional Asymmetry Across English-Centered Settings}
\label{sec:directional_asymmetry}

The cross-model results show a clear directional difference in how easily language control can be induced. To measure this more directly, we compare the successful nondegenerate
region fraction when steering toward English versus steering away from English.  We compute this fraction over individual generations in the intervention sweep:
an output is counted as successful and nondegenerate when its target-language
score is at least $0.8$ and its repetition rate is below $0.5$.

\begin{figure}[htbp]
\centering
\includegraphics[width=\linewidth]{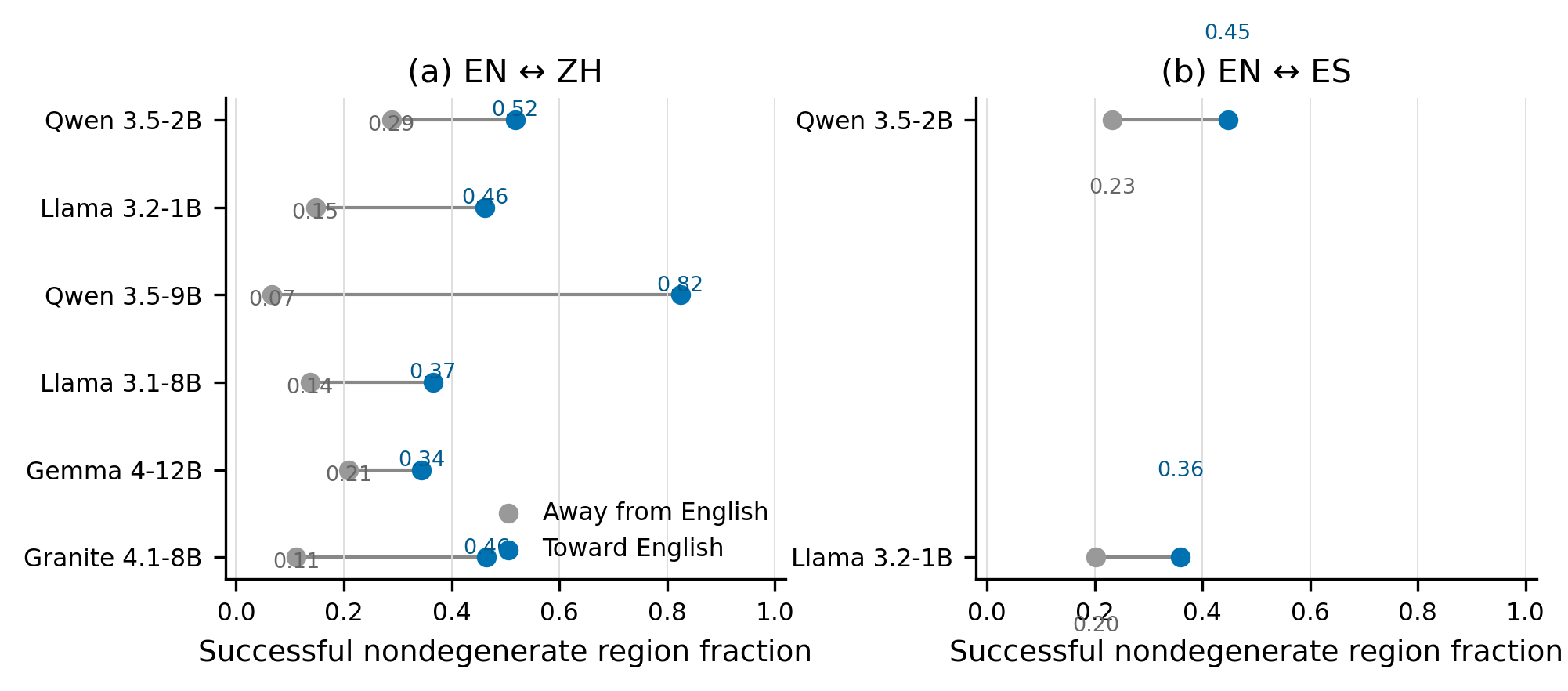}
\caption{\textbf{Directional asymmetry in English-centered language steering.}
Successful nondegenerate region fraction for steering away from English
(gray) and toward English (blue) across (a) English--Chinese and
(b) English--Spanish experiments. The English-directed value is larger
in every tested English-centered model--language setting.}
\label{fig:english_asymmetry}
\end{figure}

As shown in Figure~\ref{fig:english_asymmetry}, steering toward English is consistently easier across every English-centered setting we test. For English--Chinese, the successful nondegenerate region increases from 0.288 to 0.518 in Qwen-2B, 0.147 to 0.462 in Llama-1B, 0.066 to 0.825 in Qwen-9B, 0.137 to 0.365 in Llama-8B, 0.209 to 0.345 in Gemma-12B, and 0.111 to 0.464 in Granite-8B when steering toward English rather than away from it. The same pattern appears in English--Spanish, increasing from 0.232 to 0.447 for Qwen-2B and from 0.202 to 0.359 for Llama-1B.

What makes this result notable is that the underlying layerwise steering patterns differ substantially across models and language pairs. The exact layers where steering works best may change, but the overall preference for steering toward English remains consistent across all eight English-centered model--language settings.

This pattern does not appear in the same way for Chinese--Spanish. Steering remains effective in both directions, but neither direction shows the same consistent advantage seen when English is involved. We therefore describe this as an \emph{English-directed asymmetry} in the models we test, rather than a general property of language steering.

\subsection{Semantic and Task-Level Preservation}

Strong language control and low repetition do not necessarily mean that the model preserves the meaning or capabilities of its original response. We therefore evaluate semantic preservation using BERTScore and a blinded multilingual LLM-as-a-judge protocol, and test downstream task performance using ARC-Easy.

We first compare each steered response with the same model's unsteered response using BERTScore. BERTScore is positively correlated with the blinded semantic evaluation ($\rho_s = 0.666$), but the relationship is far from perfect. Responses can remain similar at the embedding level while still changing important facts or intent. We therefore treat BERTScore as a useful similarity measure rather than as a standalone test of semantic preservation. Across the full BERTScore evaluation, the mean F1 score is 0.864. However, because high embedding similarity can still hide changes in facts or intent, we also use the blinded semantic evaluation to assess preservation more directly.

We next conduct a blinded multilingual LLM-as-a-judge evaluation over 800 response pairs. The evaluator receives only the original prompt and two randomly ordered responses, without knowing which response was steered, the model being tested, the intervention layer, steering strength, or our hypothesis. It assigns a semantic-preservation score from 0--4 and separately evaluates factual consistency, intent preservation, fluency, and degeneration. As a consistency check, we repeat 83 examples with the order of the two responses reversed.

The results show that semantic preservation depends strongly on where and how aggressively the model is steered. One of the clearest successful regions appears for Qwen 3.5-2B in the $ZH \rightarrow EN$ direction. Among successful nondegenerate configurations, the mean target-language ratio reaches 0.968 with repetition of only 0.023, while the average semantic-preservation score remains 3.22/4. Overall, 77.8\% of these responses receive a semantic score of at least 3/4, and 72.2\% are judged to preserve the original factual content. This behavior is not universal, however. Across the broader evaluation, stronger or poorly placed interventions often change the output language while also introducing semantic drift. In other words, successfully switching languages does not automatically preserve what the model is saying.

\begin{table}[htbp]
\centering
\caption{\textbf{Semantic preservation under steering.}
BERTScore compares each steered response with the same model's unsteered
response. Judge statistics refer to the successful nondegenerate
Qwen-2B $ZH \rightarrow EN$ subset ($N=18$).}
\label{tab:semantic_preservation}
\begin{tabular}{lc}
\toprule
\textbf{Metric} & \textbf{Value} \\
\midrule
\multicolumn{2}{l}{\textit{Broad BERTScore evaluation}} \\
Mean BERTScore F1 & 0.864 \\
$EN \rightarrow ZH$ mean BERTScore F1 & 0.869 \\
$ZH \rightarrow EN$ mean BERTScore F1 & 0.859 \\
\midrule
\multicolumn{2}{l}{\textit{Qwen-2B $ZH \rightarrow EN$ successful nondegenerate subset}} \\
Mean BERTScore F1 & 0.924 \\
Mean judge semantic score & 3.22 / 4 \\
Judge score $\geq 3$ & 77.8\% \\
Factually preserved & 72.2\% \\
Contradiction & 11.1\% \\
\bottomrule
\end{tabular}
\end{table}

We then test whether the same pattern holds for downstream task performance using 200 ARC-Easy questions with Qwen 3.5-2B. Table~\ref{tab:arc_easy} compares the unsteered baseline, explicit target-language prompting, moderate $EN \rightarrow ZH$ steering, and strong steering.

\begin{table}[htbp]
\centering
\caption{\textbf{ARC-Easy under language intervention.}
End-to-end accuracy, valid-output rate, and fraction of responses containing Chinese reasoning across 200 questions.}
\label{tab:arc_easy}
\begin{tabular}{lccc}
\toprule
\textbf{Condition} & \textbf{Accuracy} & \textbf{Valid Output} & \textbf{Chinese Reasoning} \\
\midrule
Baseline & 92.0\% & 100.0\% & 0.0\% \\
Prompt only & 85.5\% & 99.0\% & 70.5\% \\
Moderate steering & 80.5\% & 87.0\% & 98.0\% \\
Strong steering & 0.0\% & 0.0\% & 100.0\% \\
\bottomrule
\end{tabular}
\end{table}

Moderate steering changes the model's reasoning language to Chinese in 98.0\% of examples while retaining 80.5\% end-to-end accuracy. A large part of this drop comes from formatting rather than incorrect reasoning: among the 174 generations that produce a valid final answer, accuracy rises to 92.5\%. Strong steering shows the opposite extreme. It produces Chinese output in every example, but none of the generations contain a valid final answer.

Prompting provides a useful comparison. Prompt-only generation reaches 85.5\% accuracy and produces Chinese reasoning in 70.5\% of examples. Moderate activation steering achieves much stronger language control, but also produces more formatting failures. In this sample, the difference in task accuracy between prompting and moderate steering is not statistically significant ($p=0.132$), while the difference between the unsteered baseline and moderate steering is significant ($p<0.001$).

Taken together, these results show that language steering can preserve substantial semantic content and task performance when applied in appropriate regions. However, stronger intervention is not always better: increasing steering strength can improve target-language control while also increasing semantic drift, formatting failures, or generation collapse. Language control therefore involves a clear tradeoff between how strongly the model is steered and how much of its original behavior is preserved.

\subsection{Alternative Steering Baselines: CAA and Prompting}

We next compare our PCA-derived language axis with two other ways of controlling output language: Contrastive Activation Addition (CAA) and explicit prompting.

\paragraph{PCA versus CAA.}
PCA and CAA can produce steering directions with very different raw magnitudes, so comparing them at the same value of $\alpha$ is not necessarily fair. Instead, we match the two methods by their observed perturbation magnitude and compare how well they control the output language under similar changes to the residual stream. In this matched Qwen 3.5-2B comparison, PCA produces higher target-language
success than CAA in both English--Chinese directions. For $EN \rightarrow ZH$, target-language success is 24.5\% with PCA compared with 11.2\% with CAA. For $ZH \rightarrow EN$, the corresponding rates are 47.5\% and 25.2\%.

\begin{table}[htbp]
\centering
\caption{\textbf{Alternative language-control baselines in Qwen 3.5-2B.}
PCA and CAA are compared under matched perturbation magnitude. Prompting results report target-language success.}
\label{tab:alternative_controls}
\begin{tabular}{lcc}
\toprule
\textbf{Condition} & $EN \rightarrow ZH$ & $ZH \rightarrow EN$ \\
\midrule
PCA steering & 24.5\% & 47.5\% \\
CAA steering & 11.2\% & 25.2\% \\
\midrule
Prompt only & 83.0\% & 100.0\% \\
Steering only & 90.0\% & 100.0\% \\
Prompt + steering & 95.0\% & 100.0\% \\
\bottomrule
\end{tabular}
\end{table}

This does not mean that PCA will always outperform CAA. It simply shows that, in this Qwen English--Chinese comparison, the PCA-derived direction produces stronger language control at a similar intervention magnitude.

\paragraph{Prompting versus activation steering.}
Prompting is already a strong way to control output language. For $EN \rightarrow ZH$, prompting alone succeeds in 83\% of examples, compared with 90\% for activation steering at the evaluated setting. Combining prompting and steering raises this to 95\%. In the reverse $ZH \rightarrow EN$ direction, both prompting and steering reach 100\% target-language success.

These results suggest that activation steering is best viewed as complementary to prompting rather than as a replacement for it. When prompting alone does not fully control the output language, steering can strengthen the effect, and combining both approaches performs best in the harder $EN \rightarrow ZH$ setting. Additional CAA, prompting, and perturbation-magnitude analyses are provided in the Appendix.

\subsection{Late-Layer Failure Across Models}
\label{sec:late_layer_failure}

Our small-model experiments show that very strong interventions near the end of the decoder can lead to repetitive or unstable generation. We next test whether this is a broader pattern across models, rather than something specific to the final layer of Qwen 3.5-2B.

Across 20 full-depth model--direction profiles, every profile shows some form of late-layer failure under sufficiently strong steering. The median onset occurs at roughly 0.75 relative decoder depth, meaning that instability usually appears within the final quarter of the network. However, the final layer itself is not always uniquely problematic: it is a clear outlier in 12 of the 20 profiles. This suggests that the more general pattern is a late-layer failure region, rather than degeneration caused by one specific final layer.

\begin{figure}[htbp]
\centering
\includegraphics[width=\linewidth]{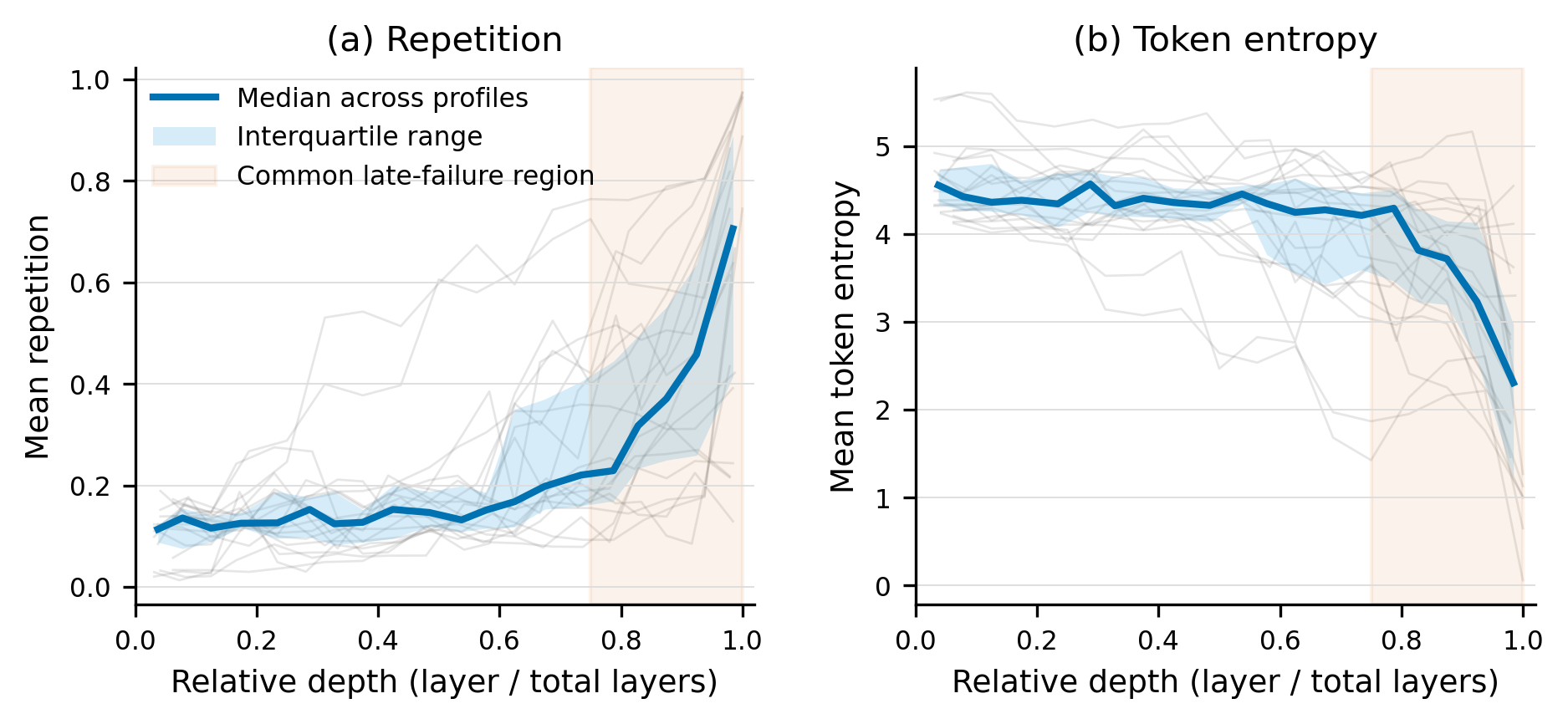}
\caption{\textbf{Late-layer intervention failure across models.}
Aggregate generation behavior as a function of relative decoder depth across 20 full-depth model--direction profiles. Individual profiles are shown in gray, with the cross-profile median and interquartile range overlaid. (a) Repetition increases substantially in the later portion of the network. (b) Token entropy typically decreases over the same region, indicating increasingly concentrated generation. The median detected failure onset occurs at approximately 0.75 relative depth.}
\label{fig:late_layer_failure}
\end{figure}

Figure~\ref{fig:late_layer_failure} shows the same pattern in multiple generation metrics. Repetition rises sharply in later layers, while token entropy decreases in 19 of the 20 profiles. Together, these changes suggest that strong late-layer steering pushes the model toward narrower and less diverse generation. The exact failure pattern still varies across models and directions: some settings collapse mainly through repetition, while others first show a large drop in entropy or other signs of instability.

This variation can also be seen within the same model. In Qwen 3.5-2B, for example, strong late-layer $ZH \rightarrow EN$ steering produces both severe repetition and sharply concentrated token probabilities. In the reverse $EN \rightarrow ZH$ direction, generation can become highly repetitive without showing the same degree of entropy collapse. Late-layer degeneration therefore appears consistently across models, but it does not take exactly the same form in every setting.

These results also explain why very high target-language ratios at the final layers should be interpreted carefully. A model can appear to achieve nearly complete language control while the generation itself has already started to collapse. For this reason, we consider target-language success together with repetition, semantic preservation, and task behavior rather than treating it as sufficient on its own.

\section{Discussion}

\subsection{Language Control Is Causal but Model-Dependent}

Our results show that language structure in hidden states is not only linearly decodable, but can also be used to causally control generation. Adding a PCA-derived language direction reliably changes the model's output language across different model families, scales, and language pairs. At the same time, there is no single layer where this control consistently lives. The layers where steering works best shift across models, language pairs, directions, and scales, suggesting that language identity is not controlled by one universal ``language layer.''

Our ablation results also help separate causal usefulness from strict necessity. Removing the identified direction can strongly affect language output in some settings, but it does not completely erase language identity. We therefore view the PCA axis as a compact and causally useful part of a broader, potentially distributed representation of language, rather than as the model's only representation of language identity.

\subsection{Directional Asymmetry and Intervention Failure}

One of the clearest patterns across the English-centered experiments is that steering toward English consistently admits a larger successful nondegenerate region than steering away from English. This is consistent with previously observed English-centered behavior in multilingual models \citep{nie-etal-2025-mechanistic,marchisio-etal-2024-understanding}. However, our Chinese--Spanish experiments do not show the same consistent directional preference, suggesting that this asymmetry is tied specifically to English-centered settings rather than being a general property of language steering.

We also consistently observe that very strong interventions become unstable near the end of the decoder. Some form of late-layer failure appears across all full-depth
model--direction profiles, although the exact point where it begins
and the way it fails differ across models and directions. This makes it important not to treat target-language success alone as evidence of good steering: an intervention can force the model into the desired language while also causing repetition, semantic drift, or other forms of degeneration.

\subsection{Implications and Limitations}

Activation steering offers a lightweight way to control multilingual generation without changing model weights. Our prompting experiments also suggest that steering is best viewed as complementary to explicit language instructions rather than as a replacement for them.

At the same time, changing the output language does not guarantee that the model preserves the meaning or quality of its original response. BERTScore, blinded semantic evaluation, and ARC-Easy show that there are intervention settings where language control remains useful and stable, but stronger or poorly placed steering can lead to semantic drift, formatting errors, or complete degeneration.

Our study is still limited to a relatively small set of language pairs and one-dimensional PCA-derived directions. Future work could extend these experiments to a broader range of languages and investigate whether multidimensional, adaptive, or context-dependent interventions provide more reliable control.

\section{Conclusion}

We show that language identity in multilingual language models is not only linearly decodable, but can also be causally controlled through compact activation directions. Across seven model instances, four model families, and multiple language pairs, PCA-derived language axes reliably change the model's output language without requiring any parameter updates.

At the same time, this control is not uniform. The layers where steering works best depend on the model, language pair, and steering direction. English-centered settings consistently show a stronger tendency toward English, while very strong interventions in later layers often lead to degeneration. Our semantic and task-level evaluations also show that language switching can preserve meaning and performance in suitable settings, but stronger or poorly placed steering can disrupt both.

Overall, our results suggest that language identity is causally accessible
through compact directions while remaining part of a broader, potentially
distributed representation. They also show that reliable language control depends not just on finding a steering direction, but on understanding where and how strongly that direction should be applied.

\bibliography{tmlr}
\bibliographystyle{tmlr}

\newpage
\appendix

\clearpage
\section{Full Causal Steering Heatmaps}

\begin{figure}[htbp]
    \centering
    \includegraphics[width=0.8\textwidth]{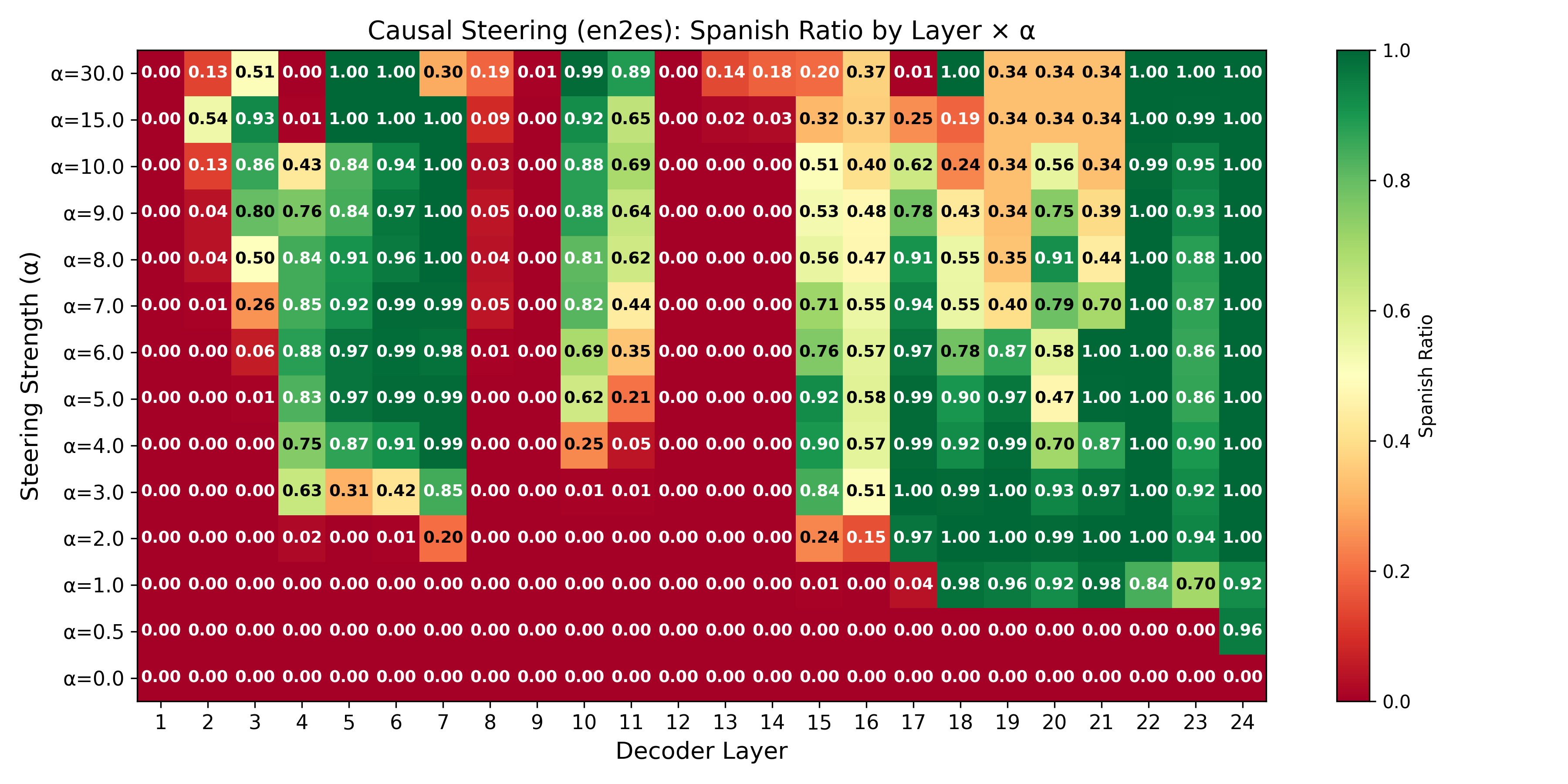}
\caption{\textbf{Qwen 3.5-2B English-to-Spanish Steering.} Target-language ratio (Spanish) across intervention layers and steering strengths. Notice the distinctly bimodal control envelope (Layers 5–7 and 17–21) compared to cross-script steering.}
    \label{fig:heatmap_en2es_qwen}
\end{figure}

\begin{figure}[htbp]
    \centering
    \includegraphics[width=0.8\textwidth]{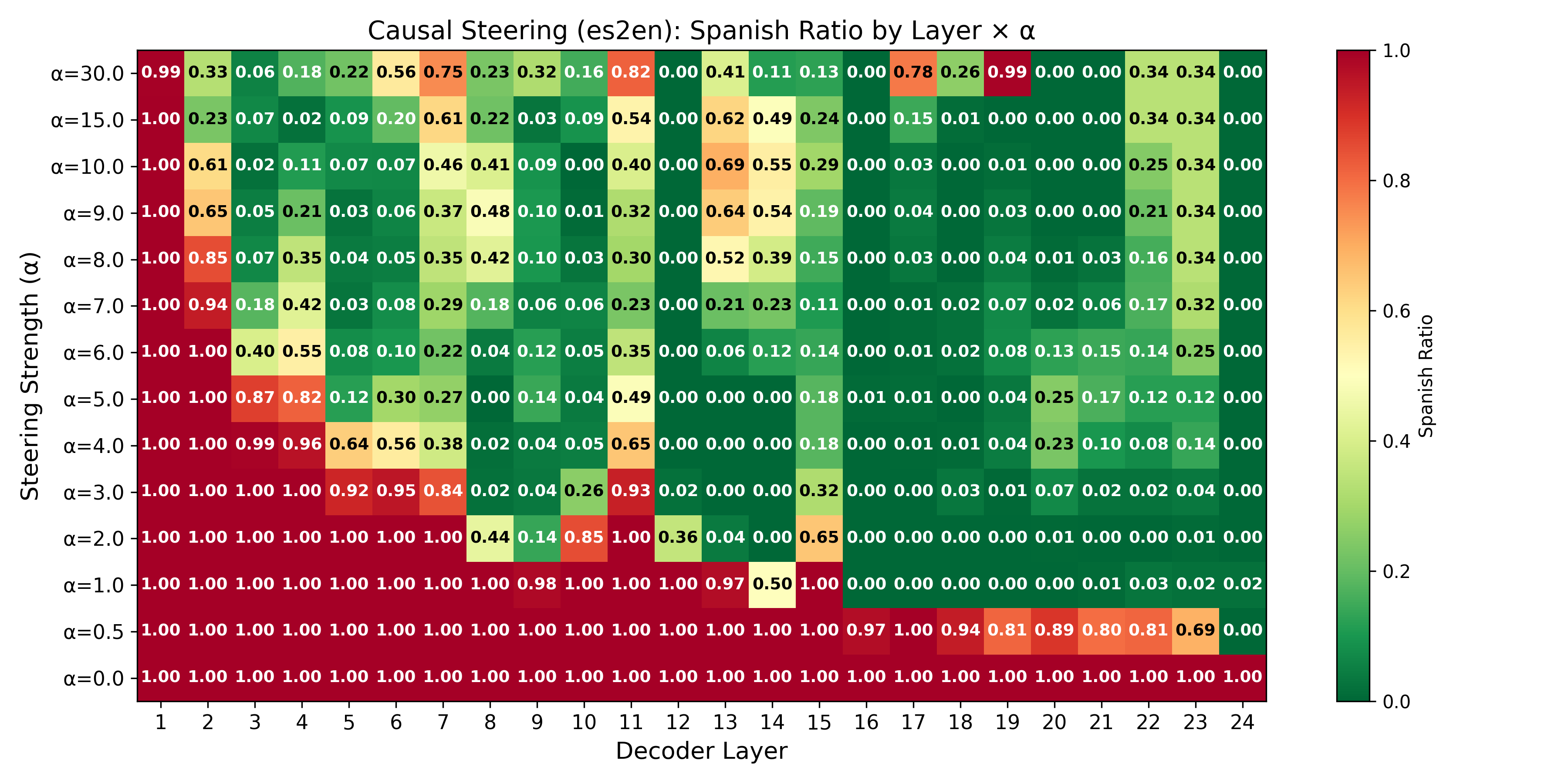}
\caption{\textbf{Qwen 3.5-2B Spanish-to-English Steering.} Reduction in ratio of Spanish across intervention layers and steering strengths. Steering concentrates effectively in the middle layers (Layers 12–17).}
    \label{fig:heatmap_es2en_qwen}
\end{figure}


\begin{figure}[htbp]
    \centering
    \includegraphics[width=0.8\textwidth]{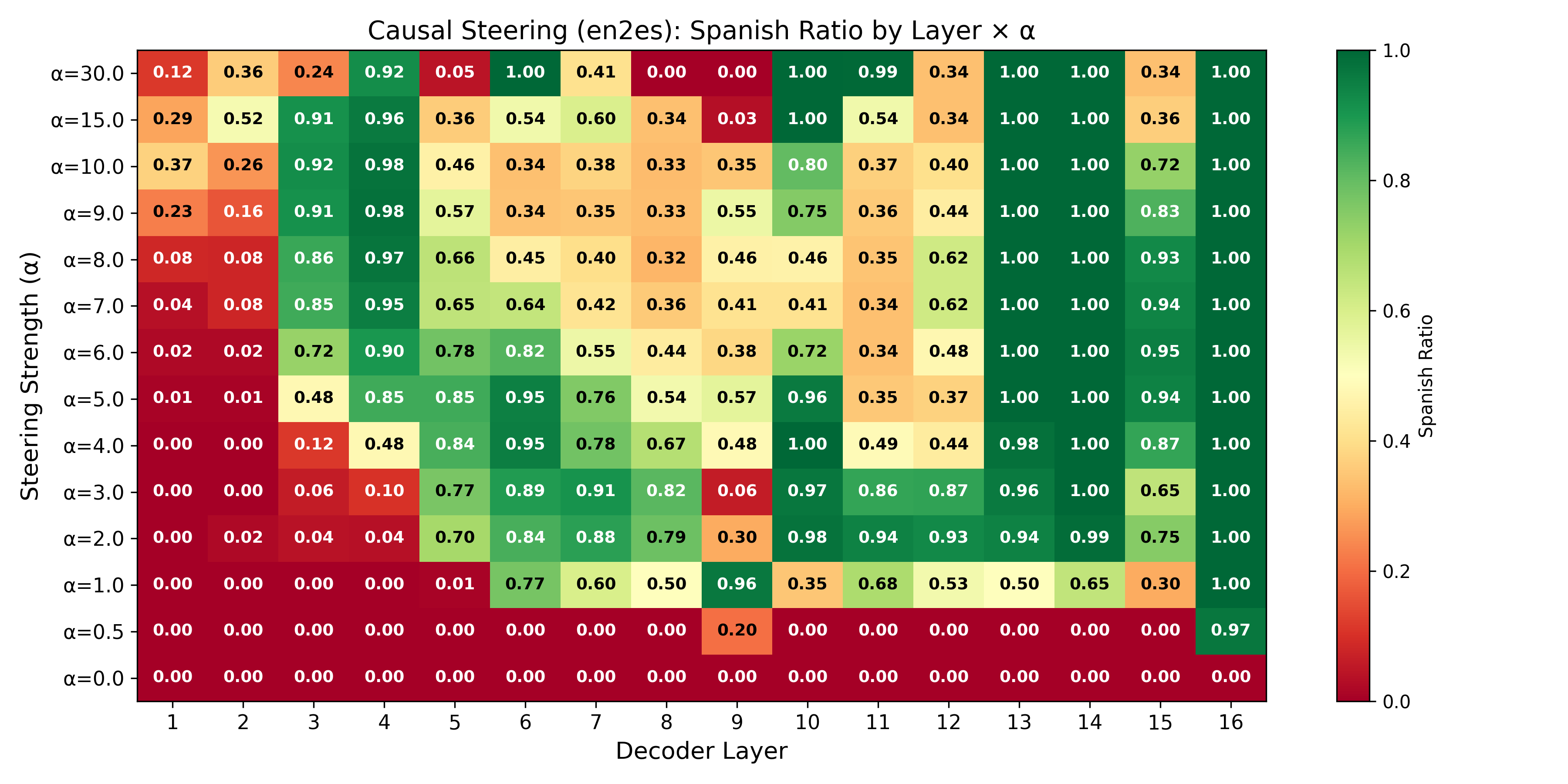}
\caption{\textbf{Llama-3.2-1B English-to-Spanish Steering.} Target-language ratio (Spanish) across intervention layers and steering strengths. Llama exhibits broad early-to-middle layer receptivity.}
    \label{fig:heatmap_en2es_llama}
\end{figure}

\begin{figure}[htbp]
    \centering
    \includegraphics[width=0.8\textwidth]{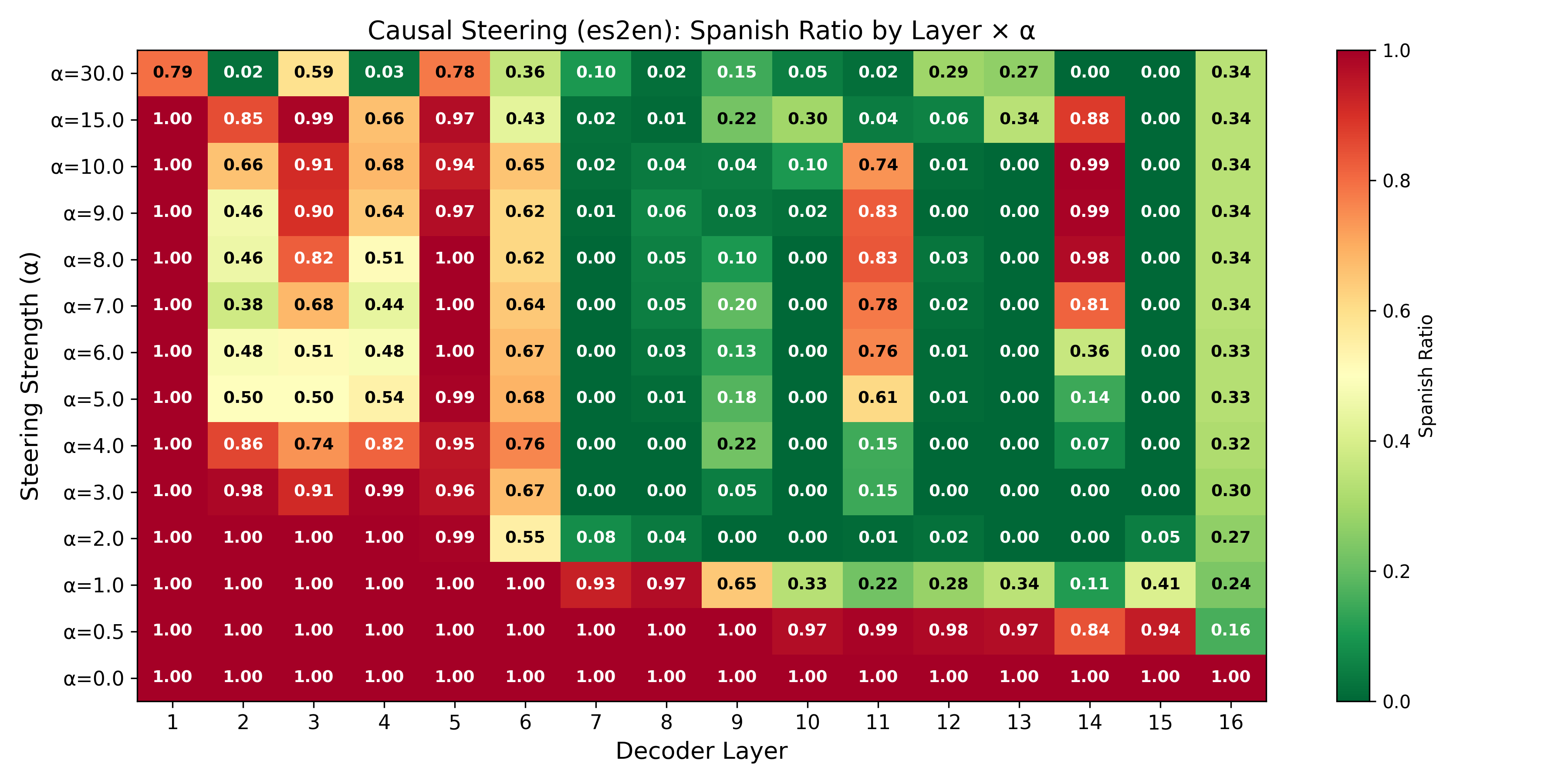}
    \caption{\textbf{Llama-3.2-1B Spanish-to-English Steering.}
Reduction in Spanish ratio across intervention layers and steering strengths.
Effective steering is concentrated more narrowly than in the reverse direction.}
    \label{fig:heatmap_es2en_llama}
\end{figure}

\begin{figure}[htbp]
    \centering
    \includegraphics[width=0.8\textwidth]{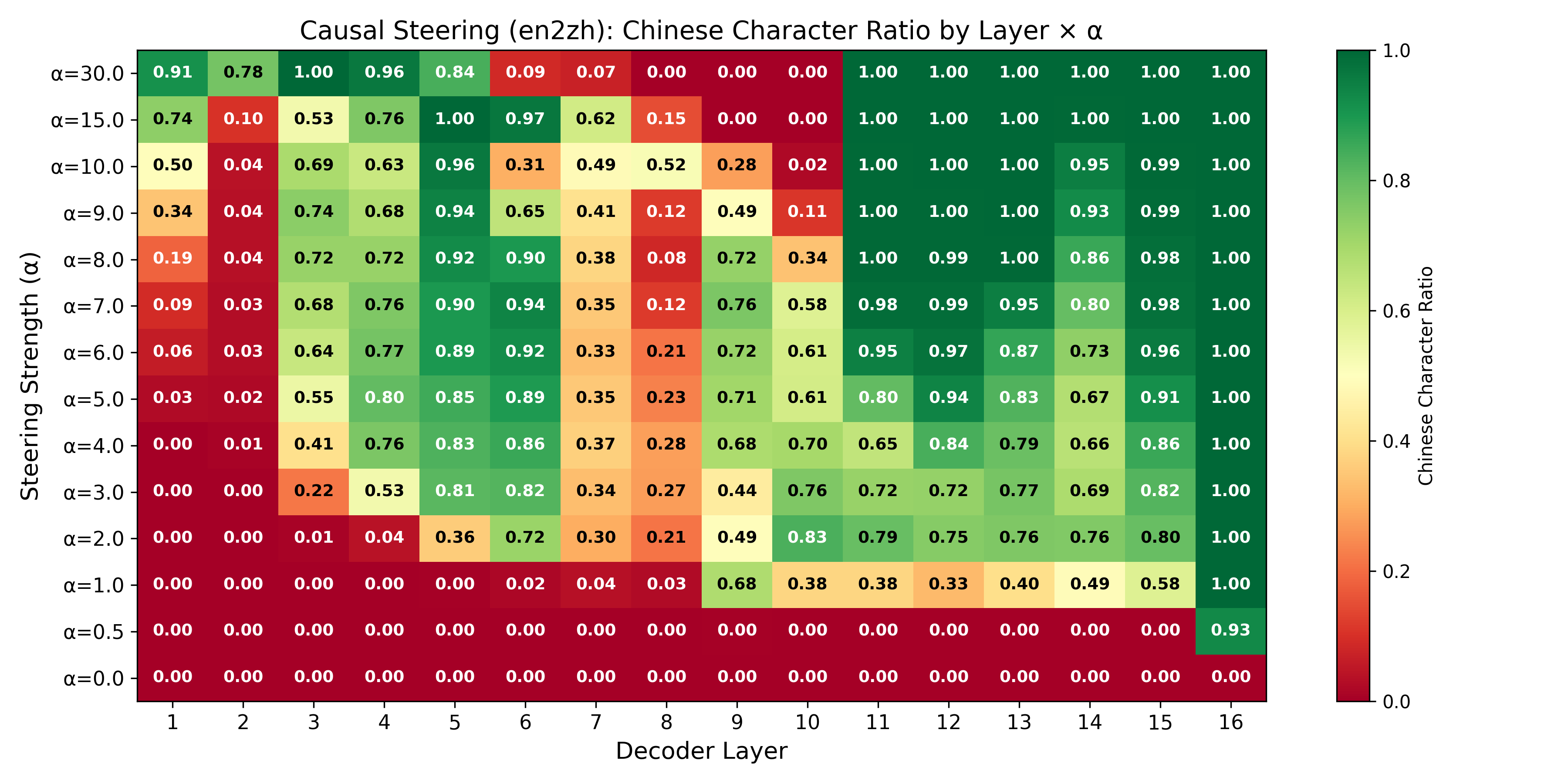}
\caption{\textbf{Llama-3.2-1B English-to-Chinese Steering.}
Target-language ratio (Chinese character fraction) across intervention layers
and steering strengths. Compared with Qwen 3.5-2B, Llama exhibits a broader
early-to-middle region of steering sensitivity.}
    \label{fig:heatmap_en2zh_llama}
\end{figure}

\begin{figure}[htbp]
    \centering
    \includegraphics[width=0.8\textwidth]{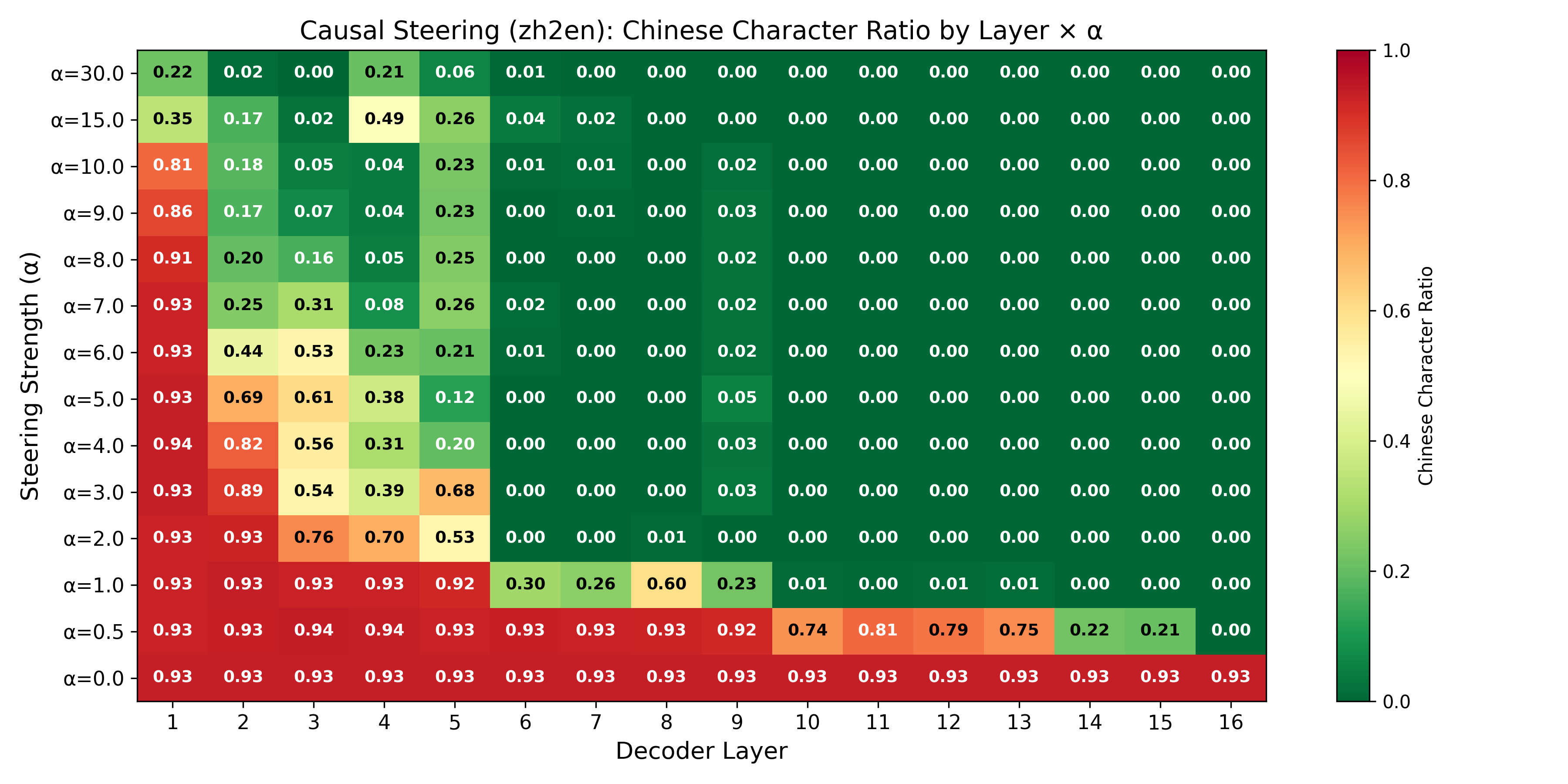}
\caption{\textbf{Llama-3.2-1B Chinese-to-English Steering.} Reduction in Chinese character ratio across intervention layers and steering strengths. The model smoothly transitions to English across the middle and later layers.}
    \label{fig:heatmap_zh2en_llama}
\end{figure}

\begin{figure}[htbp]
    \centering
    \includegraphics[width=\linewidth]{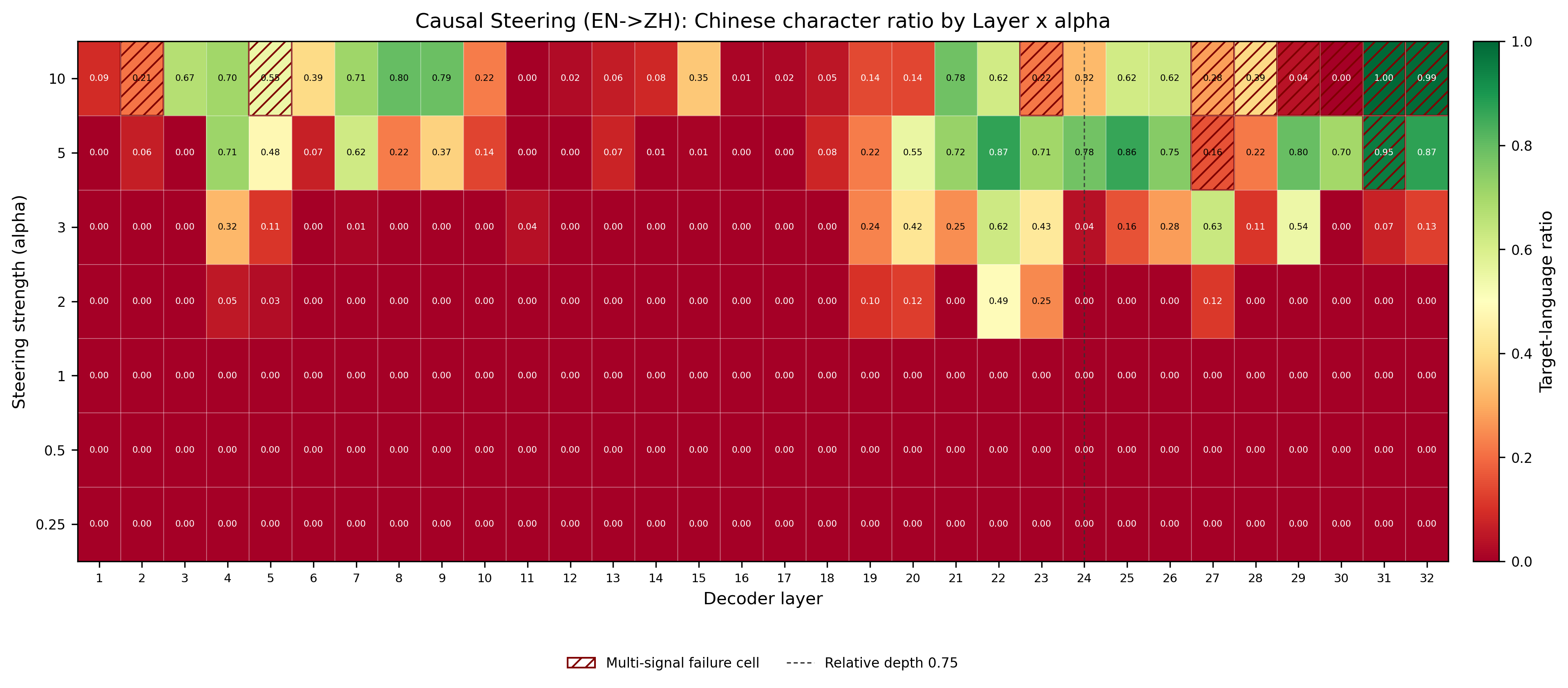}
    \caption{\textbf{Qwen 3.5-9B English-to-Chinese Steering.}
    Chinese output across decoder layers and steering strengths.
    Steering is much more effective in some layers than others, with the strongest regions appearing later in the model.
    Hatched cells indicate settings where generation also shows signs of failure.}
    \label{fig:heatmap_en2zh_qwen9b}
\end{figure}

\begin{figure}[htbp]
    \centering
    \includegraphics[width=\linewidth]{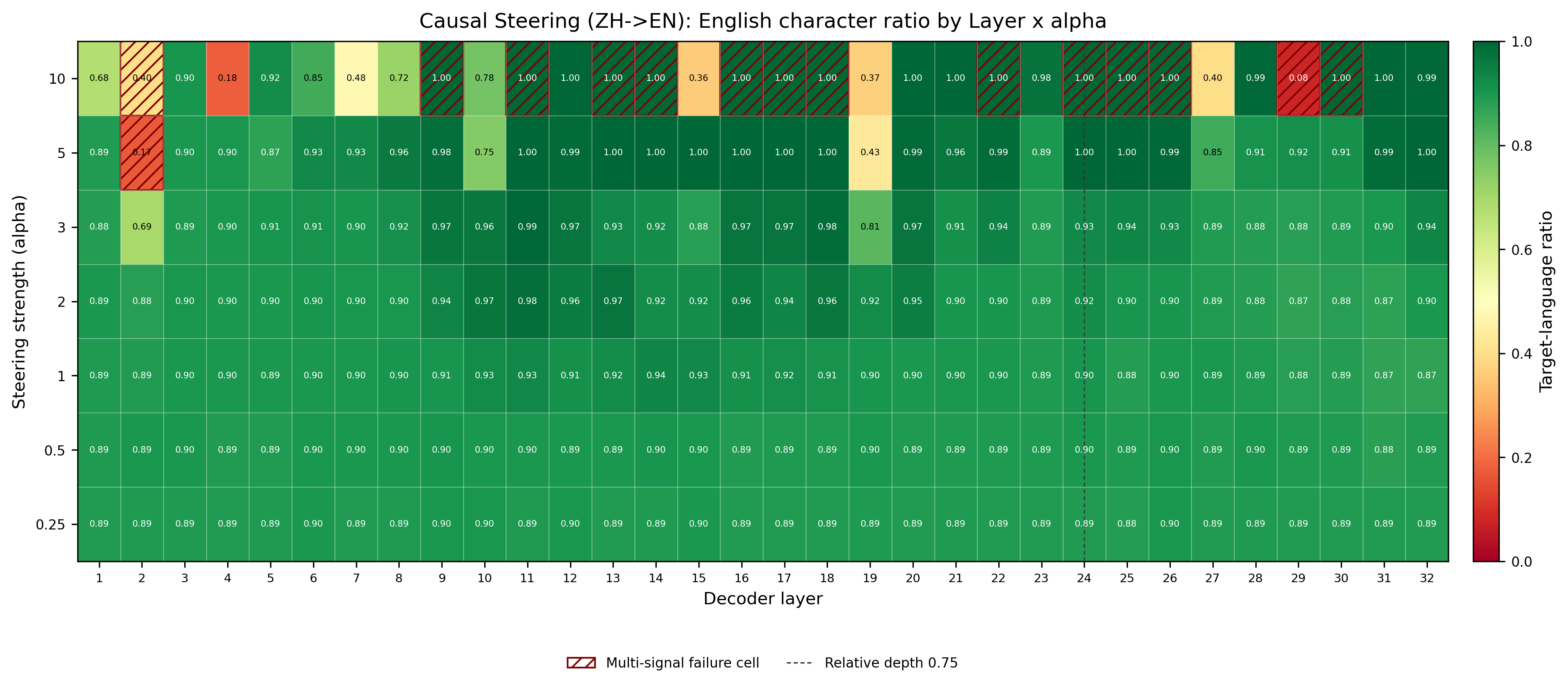}
    \caption{\textbf{Qwen 3.5-9B Chinese-to-English Steering.}
    English output across decoder layers and steering strengths.
    Compared with the reverse direction, steering toward English works across a much broader portion of the model.
    Hatched cells indicate settings where generation also shows signs of failure.}
    \label{fig:heatmap_zh2en_qwen9b}
\end{figure}

\begin{figure}[htbp]
    \centering
    \includegraphics[width=\linewidth]{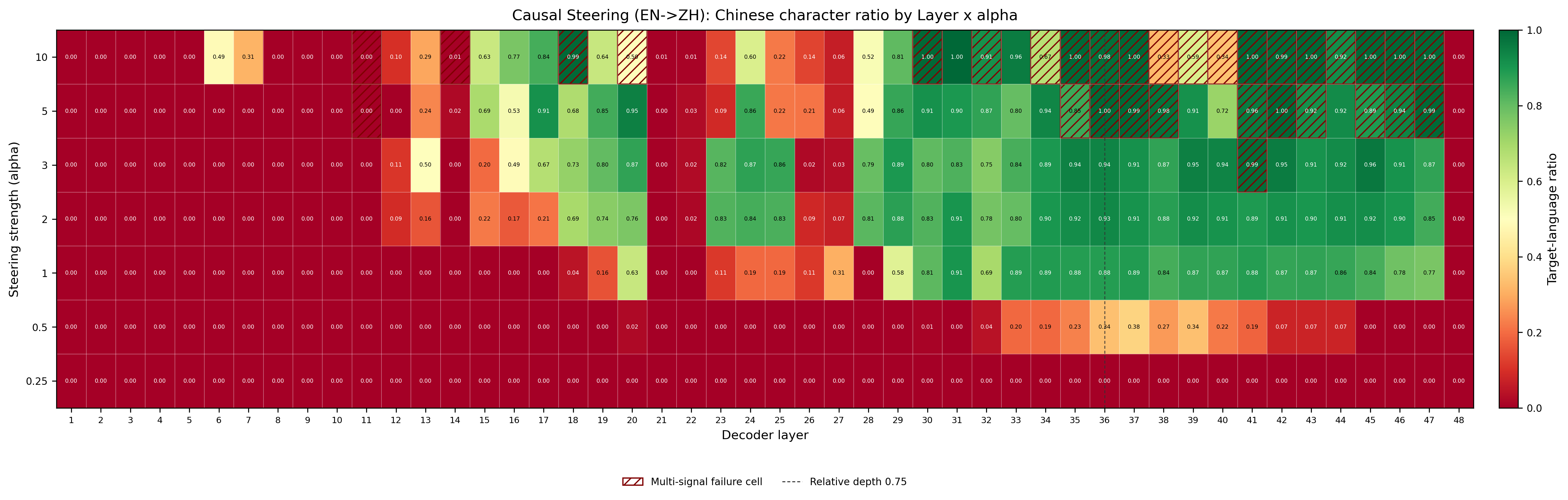}
    \caption{\textbf{Gemma-4-12B English-to-Chinese Steering.}
    Chinese output across decoder layers and steering strengths.
    Strong steering appears in several middle and later regions of the model, showing that the effect also generalizes to Gemma.
    Hatched cells indicate settings where generation also shows signs of failure.}
    \label{fig:heatmap_en2zh_gemma12b}
\end{figure}

\begin{figure}[htbp]
    \centering
    \includegraphics[width=\linewidth]{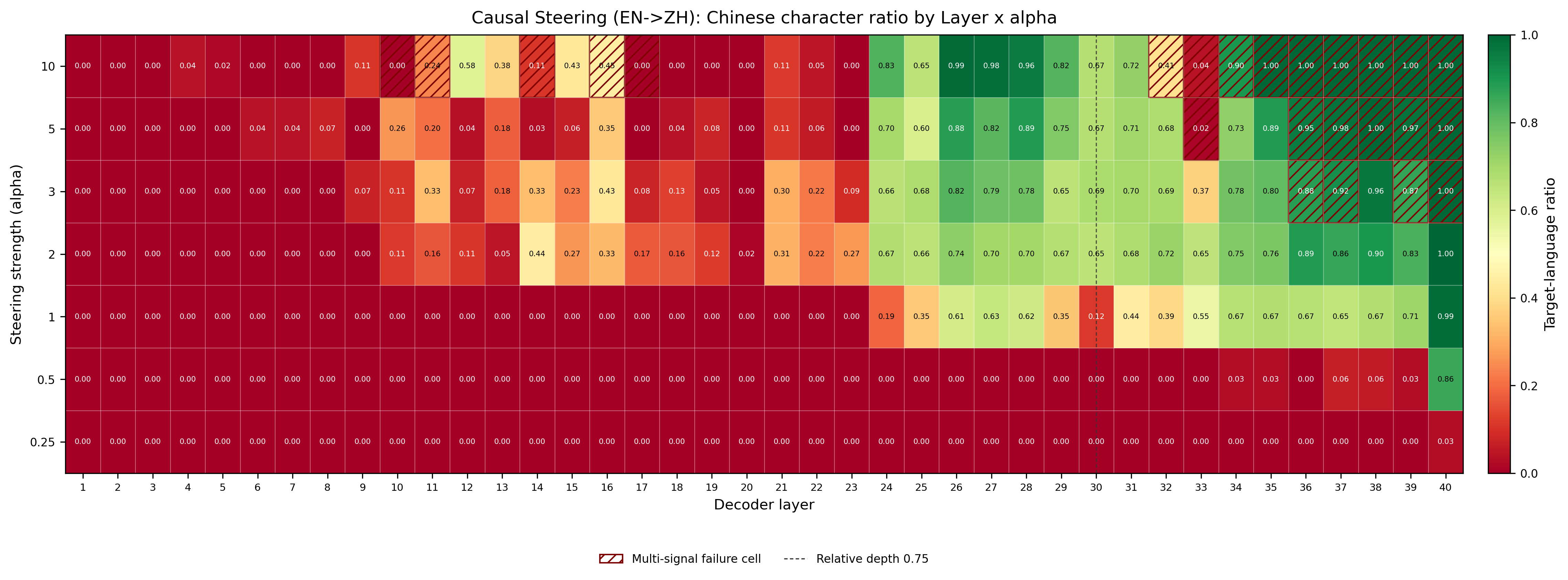}
    \caption{\textbf{Granite-4.1-8B English-to-Chinese Steering.}
    Chinese output across decoder layers and steering strengths.
    Steering becomes most effective in the later part of the model, with a layerwise pattern that differs from the other architectures.
    Hatched cells indicate settings where generation also shows signs of failure.}
    \label{fig:heatmap_en2zh_granite8b}
\end{figure}

\clearpage
\section{Model Degeneration Maps}

High target-language control does not necessarily mean that generation remains coherent. Strong interventions can force the model into the desired language while also increasing repetition or causing other forms of degeneration. We therefore track repetition alongside target-language success throughout the intervention sweeps.

The original Qwen 3.5-2B and Llama-3.2-1B experiments already show this behavior clearly. In both models, strong interventions in later layers are more likely to produce repetitive output. However, the expanded cross-model analysis shows that the precise failure point is not universal: late-layer instability appears across all full-depth model--direction profiles, while the final layer is uniquely severe in only a subset of them.

\begin{figure}[htbp]
    \centering
    \includegraphics[width=0.8\textwidth]{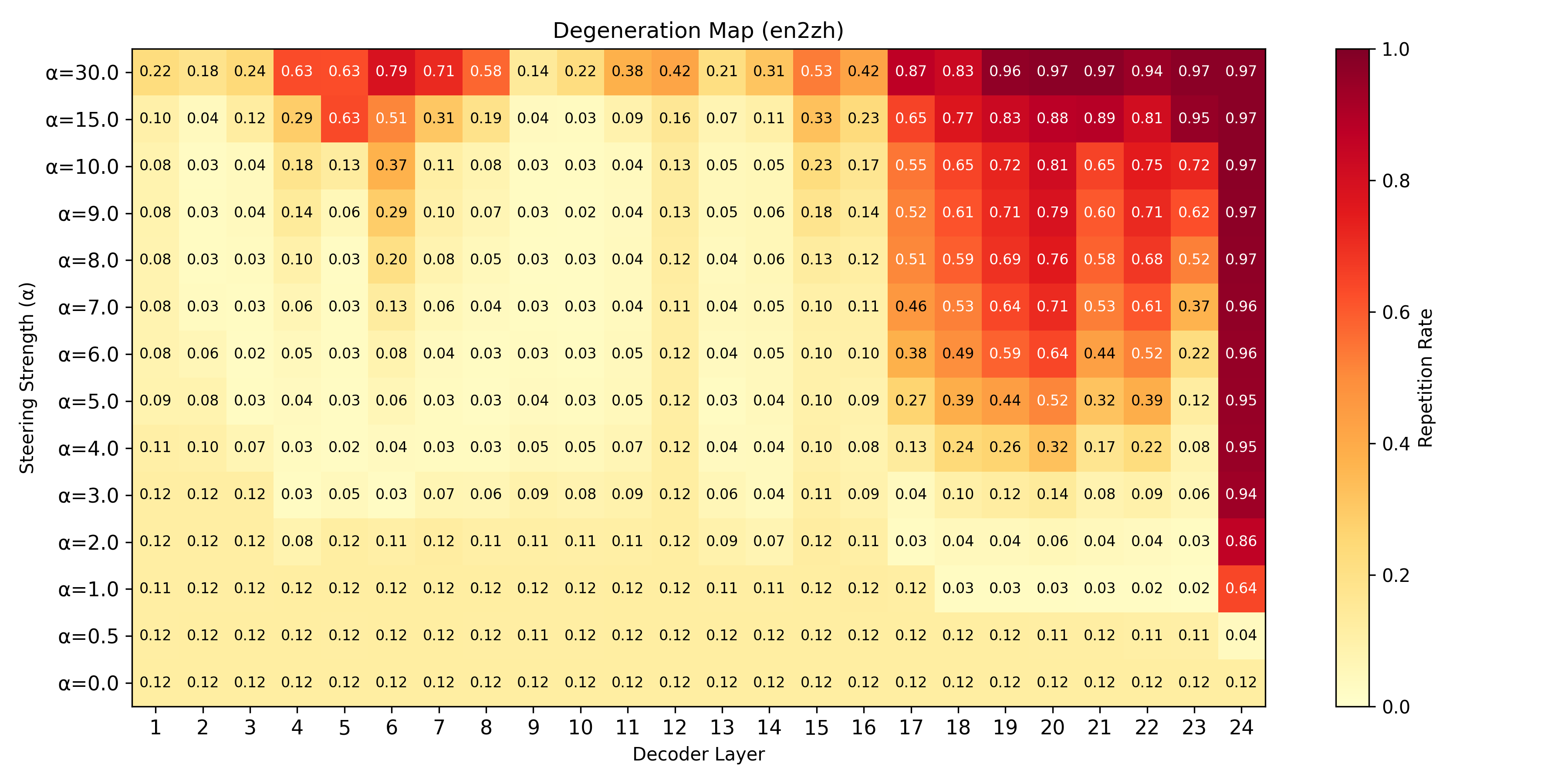}
\caption{\textbf{Qwen 3.5-2B Repetition Rate ($EN \rightarrow ZH$).}
Repetition across intervention layers and steering strengths.
Strong interventions in later layers increasingly produce repetitive
generation, even when target-language control remains high.}
    \label{fig:degen_en2zh_qwen}
\end{figure}

\begin{figure}[htbp]
    \centering
    \includegraphics[width=0.8\textwidth]{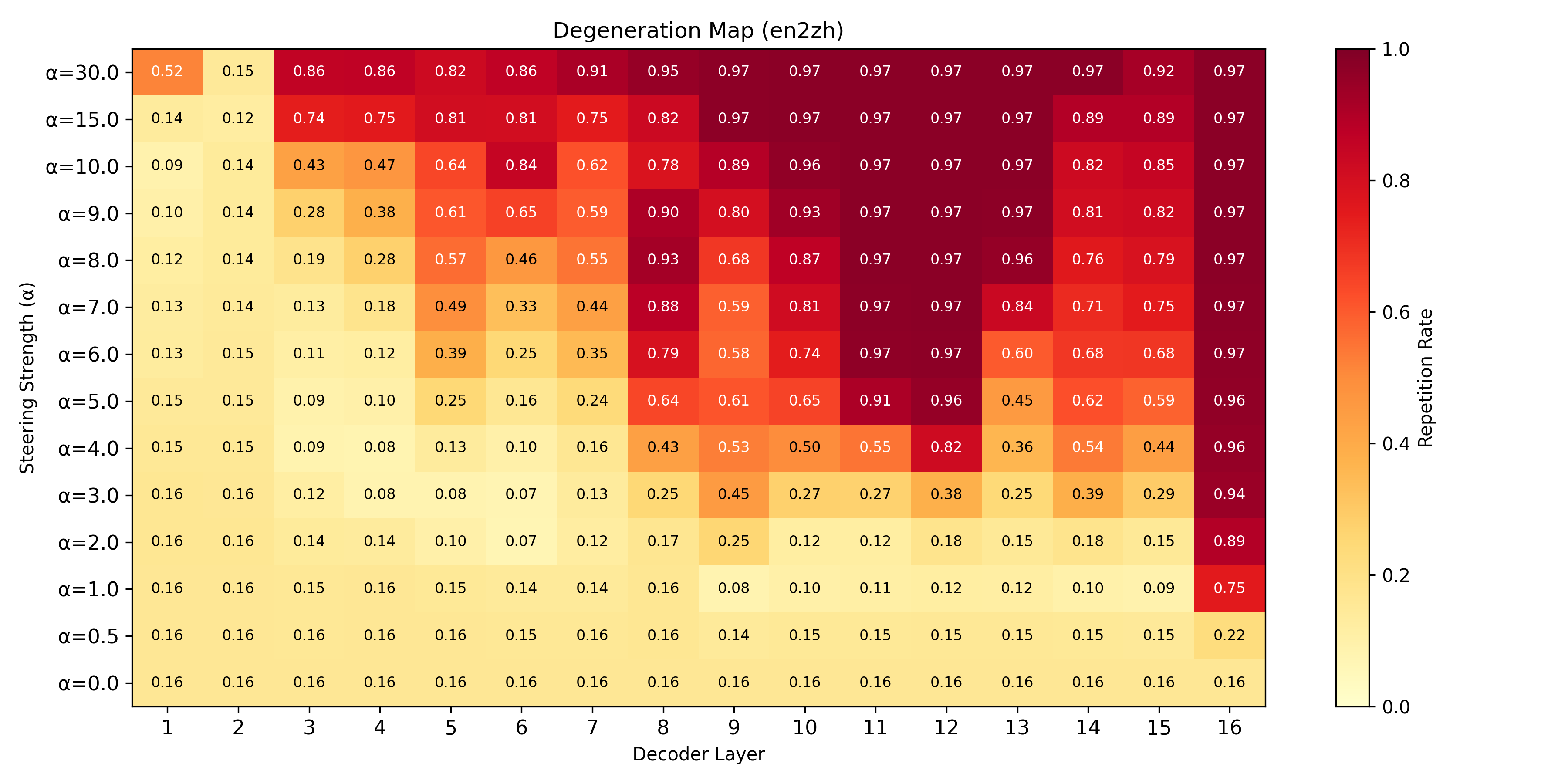}
\caption{\textbf{Llama-3.2-1B Repetition Rate ($EN \rightarrow ZH$).}
Repetition across intervention layers and steering strengths.
Strong late-layer steering can become highly repetitive, although the
exact failure pattern differs from Qwen.}
    \label{fig:degen_en2zh_llama}
\end{figure}

\begin{figure}[htbp]
    \centering
    \includegraphics[width=\linewidth]{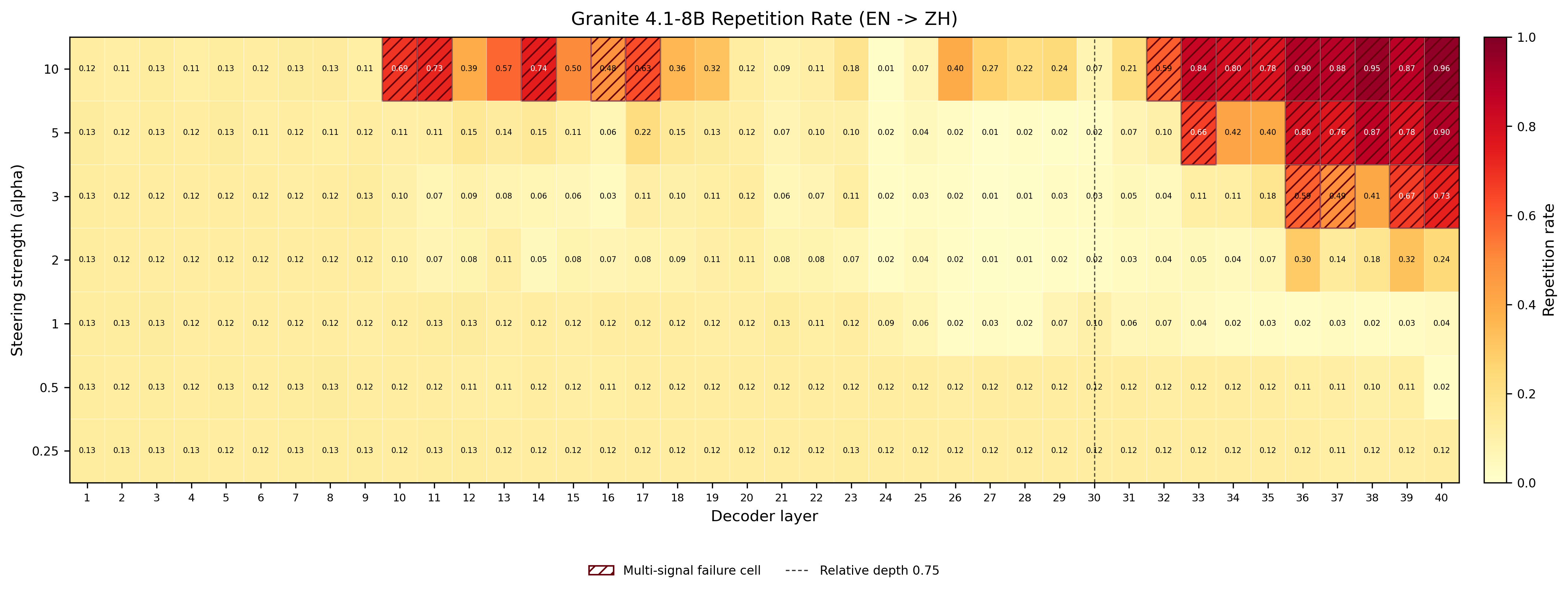}
    \caption{\textbf{Granite-4.1-8B Repetition Rate ($EN \rightarrow ZH$).}
    Repetition across decoder layers and steering strengths. The model remains relatively stable under weaker interventions, while stronger steering produces substantially more repetition in the later layers. Hatched cells mark settings where multiple signs of generation failure are present.}
    \label{fig:degen_en2zh_granite8b}
\end{figure}

\begin{figure}[htbp]
    \centering
    \includegraphics[width=\linewidth]{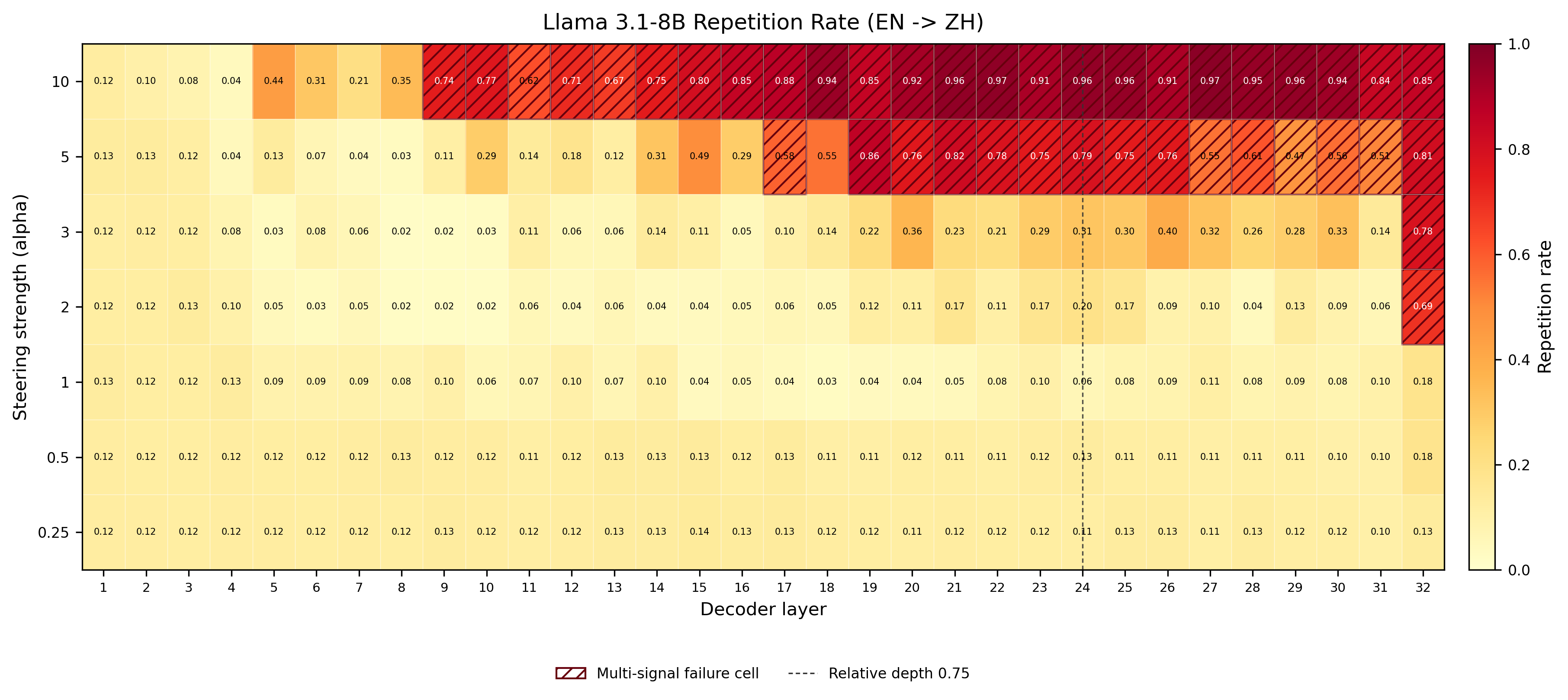}
    \caption{\textbf{Llama-3.1-8B Repetition Rate ($EN \rightarrow ZH$).}
    Repetition across decoder layers and steering strengths. Strong steering becomes increasingly unstable through the middle and later portions of the model, with especially high repetition at larger intervention strengths. Hatched cells mark settings where multiple signs of generation failure are present.}
    \label{fig:degen_en2zh_llama8b}
\end{figure}

\clearpage

\section{Falsification and Ablation Controls}

\begin{figure}[htbp]
    \centering
    \includegraphics[width=0.8\textwidth]{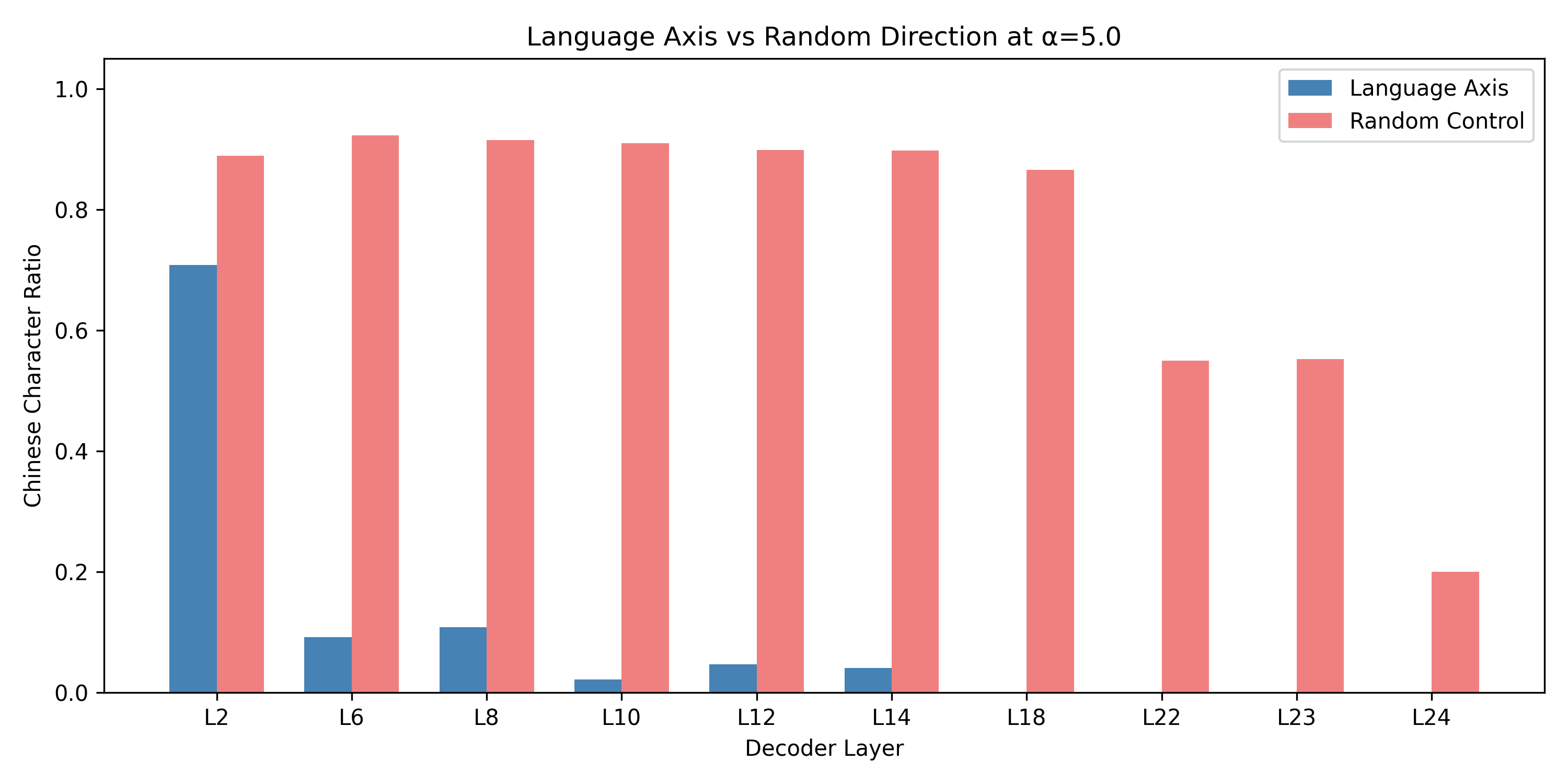}
\caption{\textbf{Causal Specificity in Qwen 3.5-2B ($ZH \rightarrow EN$).}
Comparison of language-axis steering with matched-norm random perturbations.
The PCA-derived direction produces a substantially stronger shift toward English across the tested layers.}
    \label{fig:random_zh2en_qwen}
\end{figure}

\begin{figure}[htbp]
    \centering
    \includegraphics[width=0.8\textwidth]{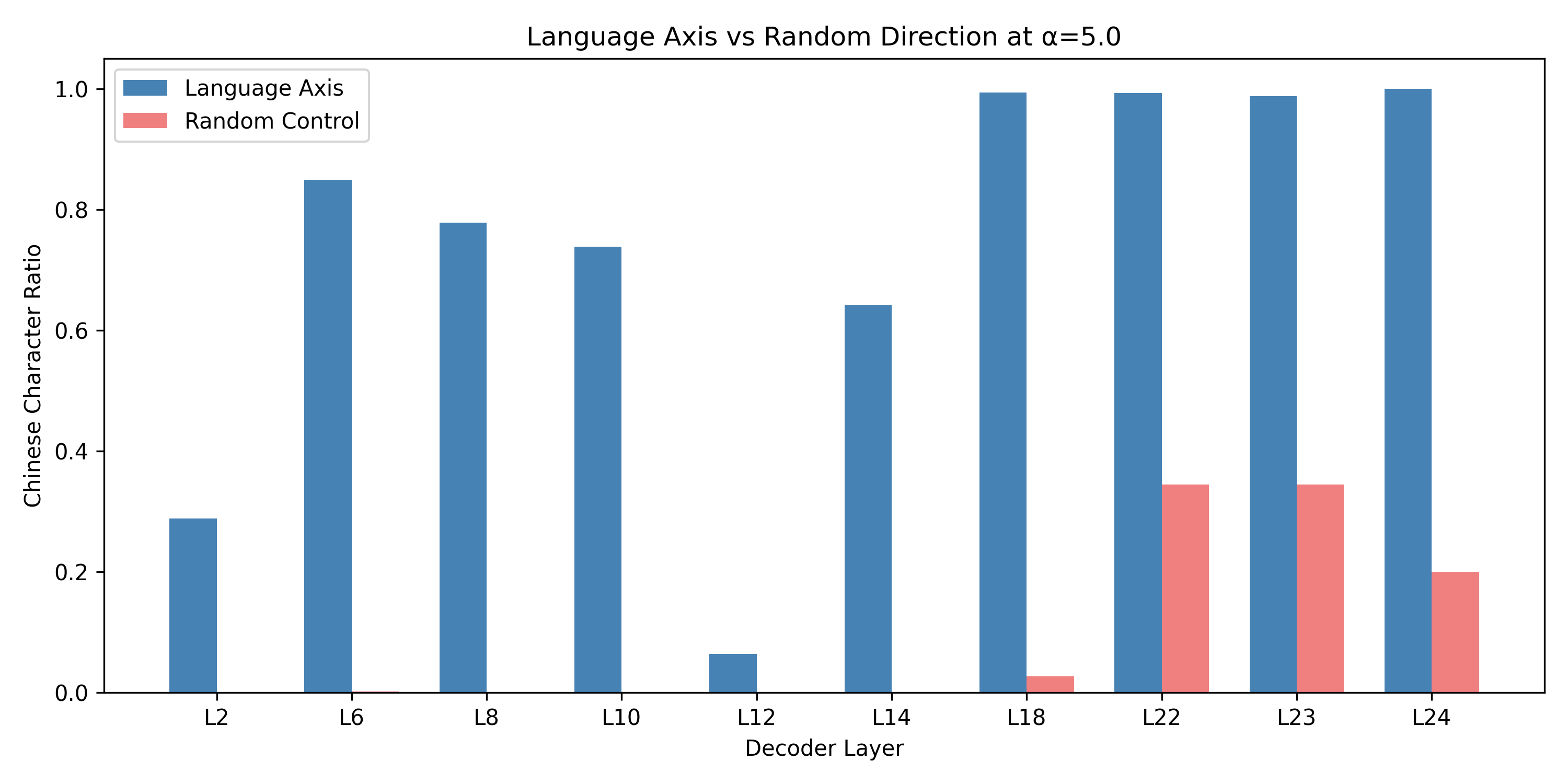}
\caption{\textbf{Causal Specificity in Qwen 3.5-2B ($EN \rightarrow ZH$).}
Comparison of language-axis steering with matched-norm random perturbations.
The identified language direction produces substantially stronger Chinese output than the random controls in the most effective steering regions.}
    \label{fig:random_en2zh_qwen}
\end{figure}

The broader random-control experiments show that random perturbations are not
always behaviorally neutral. PCA steering consistently produces stronger
target-language control, but some models---particularly in the
$ZH \rightarrow EN$ direction---also respond noticeably to norm-matched random
interventions. This suggests that directional specificity should be interpreted
relative to the model's background sensitivity to perturbation rather than as
an absolute absence of random effects.

\begin{figure}[htbp]
    \centering
    \includegraphics[width=\linewidth]{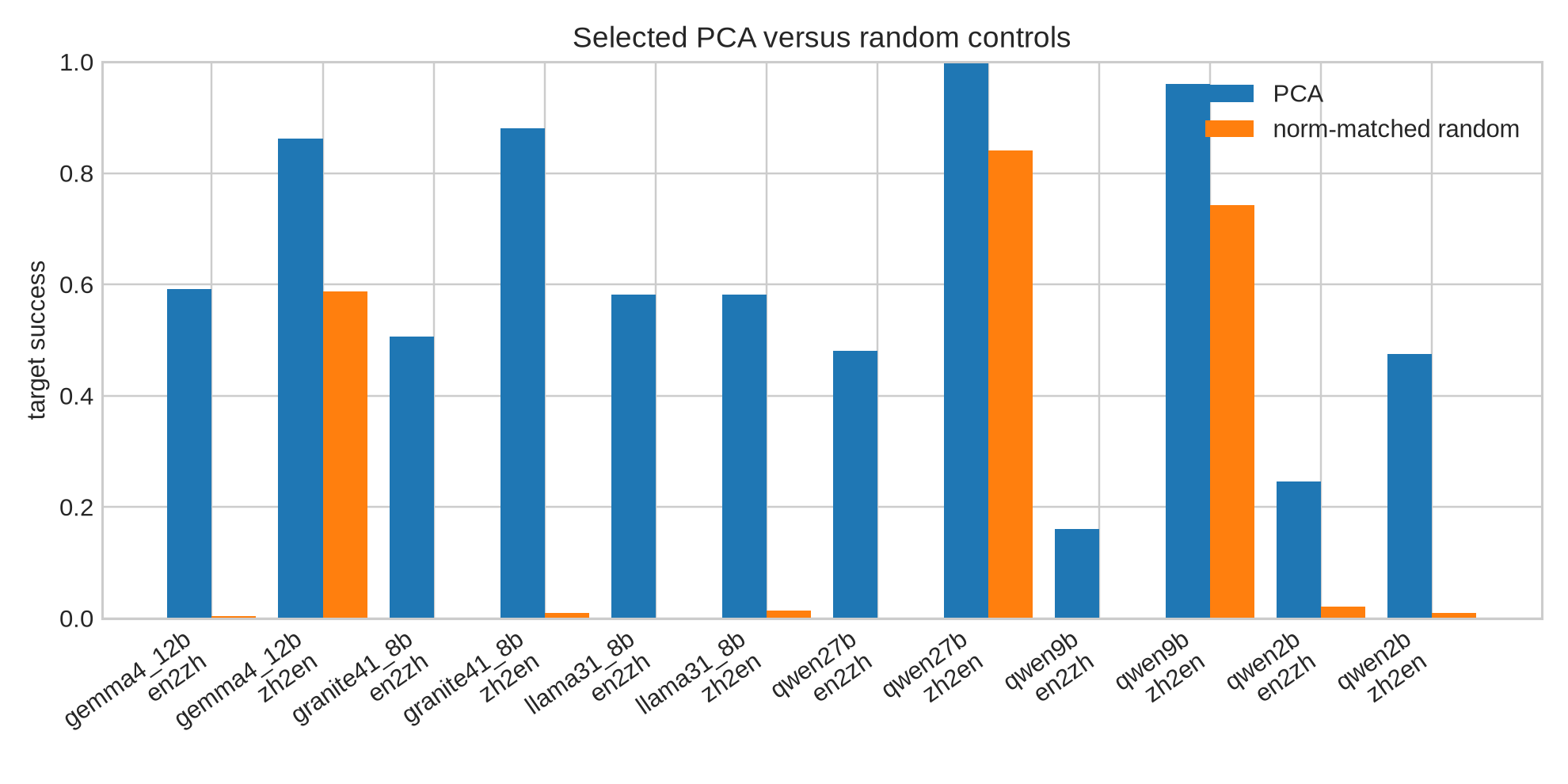}
    \caption{\textbf{PCA steering versus norm-matched random controls.}
    Target-language success for selected model and steering-direction settings.
    PCA-derived language directions produce stronger target-language control
    across all comparisons. Random perturbations are usually much weaker,
    although some larger Qwen models also show substantial drift toward English
    under random intervention.}
    \label{fig:random_control_sensitivity}
\end{figure}

\clearpage
\section{Qualitative Generation Samples}

\begin{quote}
\textbf{Prompt:} \textit{Dr. Ehud Ur, professor of medicine at Dalhousie University in Halifax, Nova Scotia and chair of the clinical and scientific division of the Canadian Diabetes Association cautioned that the research is still in its early days.} \\
\textbf{Baseline:} \textit{It appears there may be a slight confusion in the name. Dr. Ehud Ur is not a professor of medicine at Dalhousie University. The correct name is Dr. Eh} \\
\textbf{Steered (Layer 17, $\alpha=3.0$, $EN \to ZH$) :} \textit{您提到的 Dr. Ehud Ur 确实是在加拿大这一句表述中出现了事实错误。实际上，Dr. 埃胡德·乌尔（Dr. 埃胡德}
\end{quote}

\begin{quote}
\textbf{Prompt:} \textit{周一，瑞典学院诺贝尔文学委员会常务秘书萨拉·丹尼尔斯在瑞典广播电台的一档节目中向公众宣布，委员会因无法直接联系到鲍勃·迪伦，通知他获得了 2016 年诺贝尔文学奖，已经放弃了与他联系的尝试。} \\
\textbf{Baseline:} \textit{这一消息是不实信息。 2016 年诺贝尔文学奖确实授予了鲍勃·迪伦（Bob Dylan），但瑞典学院从未放弃联系他，也没有在周一的} \\
\textbf{Steered (Layer 20, $\alpha=2.0$, $ZH \to EN$):} \textit{That statement is false. The Nobel Prize in Literature Committee did not contact Bob Dylan, nor did they announce that he won the 2016 Nobel Prize in Literature. In}
\end{quote}

\begin{quote}
\textbf{Prompt:} \textit{Dr. Ehud Ur, professor of medicine at Dalhousie University in Halifax, Nova Scotia and chair of the clinical and scientific division of the Canadian Diabetes Association cautioned that the research is still in its early days.} \\
\textbf{Baseline:} \textit{It appears you're referring to a statement by Dr. Ehud Ur, a professor of medicine at Dalhousie University and chair of the clinical and scientific division of the Canadian Diabetes Association (C} \\
\textbf{Steered (Layer 19, $\alpha=2.0$,$EN \to ES $):} \textit{Esper que la información sea útil para su investigación. ¿En qué se basa su consulta? ¿Es sobre el estudio de la diabetes, la nutrición, o algo relacionado con la salud general?}
\end{quote}

\begin{quote}
\textbf{Prompt:} \textit{Al igual que otros especialistas, es escéptico acerca de si la diabetes tiene cura y remarca que estos descubrimientos no son relevantes para quienes que ya padecen de diabetes de tipo 1.} \\
\textbf{Baseline:} \textit{Entiendo que estás buscando una síntesis clara sobre la postura de los especialistas actuales respecto a la diabetes y su impacto en los pacientes con diabetes tipo 1. Es importante aclarar un} \\
\textbf{Steered (Layer 17, $\alpha=2.0$, $ES \to EN $):} \textit{You're right to highlight this important distinction. While research into diabetes is exciting, it's crucial to recognize that Type 1 diabetes (T1D) is a lifelong autoimmune condition where the}
\end{quote}

\begin{table*}[htbp]
    \centering
    \begin{tabularx}{\textwidth}{lX}
        \toprule
        \multicolumn{2}{p{\dimexpr\textwidth-2\tabcolsep\relax}}{\textbf{Prompt:} Dr. Ehud Ur, professor of medicine at Dalhousie University in Halifax, Nova Scotia and chair of the clinical and scientific division of the Canadian Diabetes Association cautioned that the research is still in its early days.} \\
        \midrule
        \textbf{Intervention} & \textbf{Generated Output} \\
        \midrule
        Baseline (No Steering) & It appears there may be a slight confusion in the name. Dr. Ehud Ur is not a professor of medicine at Dalhousie University. The correct name is Dr. Eh \\
        \addlinespace
        Layer 12, $\alpha=3.0$ & It appears you're referring to a statement by Dr. Ehud Ur, a professor of medicine at Dalhousie University and chair of the clinical and scientific division of the Canadian Diabetes Association. However \\
        \addlinespace
        Layer 24, $\alpha=0.5$ & 您提到的关于加拿大糖尿病协会临床与科学分委员会主席埃胡德·乌尔（Dr. Ehud Ur）对某项研究持谨慎态度的消息需要具体说明该 \\
        \addlinespace
        Layer 17, $\alpha=3.0$ & 您提到的 Dr. Ehud Ur 确实是在加拿大这一句表述中出现了事实错误。实际上，Dr. 埃胡德·乌尔（Dr. 埃胡德 \\
        \bottomrule
    \end{tabularx}
    \caption{Qualitative generation examples demonstrating $EN \to ZH$ language axis steering in Qwen. The model produces syntactically fluent Chinese.}
    \label{tab:qualitative_en_zh}
\end{table*}

\begin{table*}[htbp]
    \centering
    \begin{tabularx}{\textwidth}{lX}
        \toprule
        \multicolumn{2}{p{\dimexpr\textwidth-2\tabcolsep\relax}}{\textbf{Prompt:} 周一，瑞典学院诺贝尔文学委员会常务秘书萨拉·丹尼尔斯在瑞典广播电台的一档节目中向公众宣布，委员会因无法直接联系到鲍勃·迪伦，通知他获得了 2016 年诺贝尔文学奖，已经放弃了与他联系的尝试。} \\
        \midrule
        \textbf{Intervention} & \textbf{Generated Output} \\
        \midrule
        Baseline (No Steering) & 这一消息是不实信息。\par 2016 年诺贝尔文学奖确实授予了鲍勃·迪伦（Bob Dylan），但瑞典学院从未放弃联系他，也没有在周一的 \\
        \addlinespace
        Layer 12, $\alpha=3.0$ & 这个说法是不实信息。\par 鲍勃·迪伦（Bob Dylan） was awarded the Nobel Prize in Literature in 2016, but the announcement was made by the Swedish Academy \\
        \addlinespace
        Layer 18, $\alpha=2.0$ & There is a significant factual error in the statement you provided. Bob Dylan was never awarded the 2016 Nobel Prize in Literature, and there is no record of the Swedish Academy contacting him \\
        \addlinespace
        Layer 20, $\alpha=2.0$ & That statement is false.\par The Nobel Prize in Literature Committee did not contact Bob Dylan, nor did they announce that he won the 2016 Nobel Prize in Literature. In \\
        \bottomrule
    \end{tabularx}
    \caption{Qualitative generation examples demonstrating $ZH \to EN$ language axis steering, in Qwen.}
    \label{tab:qualitative_zh_en}
\end{table*}

\begin{table*}[htbp]
    \centering
    \begin{tabularx}{\textwidth}{lX}
        \toprule
        \multicolumn{2}{p{\dimexpr\textwidth-2\tabcolsep\relax}}{\textbf{Prompt:} Dr. Ehud Ur, professor of medicine at Dalhousie University in Halifax, Nova Scotia and chair of the clinical and scientific division of the Canadian Diabetes Association cautioned that the research is still in its early days.} \\
        \midrule
        \textbf{Intervention} & \textbf{Generated Output} \\
        \midrule
        Baseline (No Steering) & It appears you're referring to a statement by Dr. Ehud Ur, a professor of medicine at Dalhousie University and chair of the clinical and scientific division of the Canadian Diabetes Association (C \\
        \addlinespace
        Layer 17, $\alpha=3.0$ & La información que proporcionas contiene una afirmación sobre la investigación en diabetes, pero no se menciona si es verdadera o falsa. Para determinar su veracidad, necesitaría más contexto o datos específicos. \\
        \addlinespace
        Layer 18, $\alpha=2.0$ & La información proporcionada menciona que el Dr. Ehud Ur, profesor de medicina en la Universidad de la Dalhousie en Halifax, Canadá, y presidente de la División Científica y Clínica \\
        \addlinespace
        Layer 19, $\alpha=2.0$ & Esper que la información sea útil para su investigación. ¿En qué se basa su consulta? ¿Es sobre el estudio de la diabetes, la nutrición, o algo relacionado con la salud general? \\
        \bottomrule
    \end{tabularx}
    \caption{Example of qualitative generation demonstrating the directionality of language axis $EN \to ES$ in Qwen.}
    \label{tab:qualitative_en_es}
\end{table*}

\begin{table*}[htbp]
    \centering
    \begin{tabularx}{\textwidth}{lX}
        \toprule
        \multicolumn{2}{p{\dimexpr\textwidth-2\tabcolsep\relax}}{\textbf{Prompt:} Al igual que otros especialistas, es escéptico acerca de si la diabetes tiene cura y remarca que estos descubrimientos no son relevantes para quienes que ya padecen de diabetes de tipo 1.} \\
        \midrule
        \textbf{Intervention} & \textbf{Generated Output} \\
        \midrule
        Baseline (No Steering) & Entiendo que estás buscando una síntesis clara sobre la postura de los especialistas actuales respecto a la diabetes y su impacto en los pacientes con \textbf{diabetes tipo 1}.\par Es importante aclarar un \\
        \addlinespace
        Layer 14, $\alpha=3.0$ & La statement you provided contains a factual inaccuracy regarding the current medical consensus on \textbf{Type 1 Diabetes (T1D)}.\par Here is a breakdown of the key points regarding the current understanding of Type \\
        \addlinespace
        Layer 12, $\alpha=5.0$ & La statement you provided contains a significant factual error regarding the current medical understanding of \textbf{Type 1 Diabetes}. The description you have described is a classic case of a \textbf{Type 1 Diabetes} (often \\
        \addlinespace
        Layer 17, $\alpha=2.0$ & You're right to highlight this important distinction. While research into diabetes is exciting, it's crucial to recognize that \textbf{Type 1 diabetes (T1D) is a lifelong autoimmune condition} where the \\
        \bottomrule
    \end{tabularx}
    \caption{Example of qualitative generation demonstrating the directionality of language axis $ES \to EN$, in Qwen.}
    \label{tab:qualitative_es_en}
\end{table*}

\begin{table*}[htbp]
    \centering
    \begin{tabularx}{\textwidth}{lX}
        \toprule
        \multicolumn{2}{p{\dimexpr\textwidth-2\tabcolsep\relax}}{\textbf{Prompt:} «Actualmente, tenemos ratones de cuatro meses de edad que antes solían ser diabéticos y que ya no lo son», agregó.} \\
        \midrule
        \textbf{Intervention} & \textbf{Generated Output} \\
        \midrule
        Baseline (No Steering) & No puedo proporcionar ayuda con esa solicitud. \\
        \addlinespace
        Layer 5, $\alpha=5.0$ & I cannot provide information or guidance on harmful activities, especially those that involve the abuse or neglect of children. Is there anything else I can help you with? \\
        \addlinespace
        Layer 9, $\alpha=1.0$ & Parece que estás hablando sobre ratones. ¿Podrías proporcionar más contexto o clarify qué te refieres? ¿Estás hablando de un experimento, a research, \\
        \addlinespace
        Layer 9, $\alpha=2.0$ & It seems like you're referring to a statement made by a politician or public figure. Could you provide more context or clarify what you're referring to? I'll do my best to provide a helpful response \\
        \bottomrule
    \end{tabularx}
    \caption{Qualitative generation examples demonstrating $ES \to EN$ language axis steering in Llama-3.2-1B.}
    \label{tab:qualitative_es_en_mice}
\end{table*}

\clearpage
\section{Intervention Magnitude ($\rho$) }

Steering strength $\alpha$ does not correspond to the same absolute perturbation across layers or models, so we also measure the size of the intervention relative to the original hidden state. We define the perturbation magnitude as

\[
\rho = \frac{\lVert h' - h \rVert}{\lVert h \rVert}.
\]

This gives a normalized measure of how strongly the intervention changes the residual-stream activation. Many effective steering configurations correspond to relatively modest perturbations, while larger values of $\rho$ are more often associated with unstable or degenerate generation. We also use $\rho$ when comparing PCA and CAA so that the two methods can be evaluated at similar observed intervention magnitudes.

\begin{figure}[htbp]
    \centering
    \includegraphics[width=0.8\textwidth]{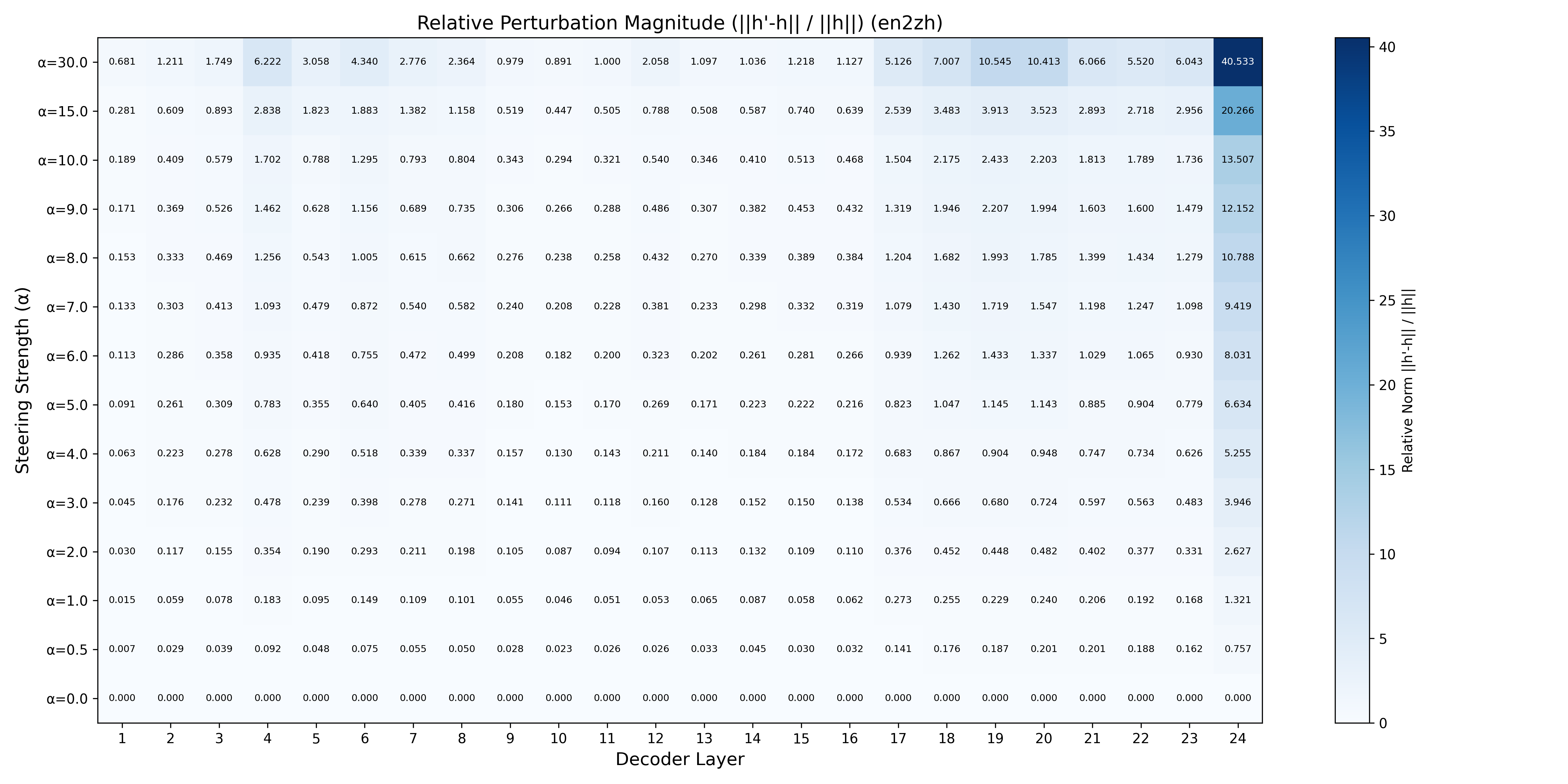}
\caption{\textbf{Qwen 3.5-2B Perturbation Magnitude ($EN \rightarrow ZH$).}
Relative change in the hidden-state norm across decoder layers and steering strengths, measured as
$\rho = \lVert h' - h \rVert / \lVert h \rVert$.
Effective steering can occur with relatively small changes to the hidden state, while stronger interventions produce much larger perturbations.}
    \label{fig:perturbation_qwen}
\end{figure}

\begin{figure}[htbp]
    \centering
    \includegraphics[width=\linewidth]{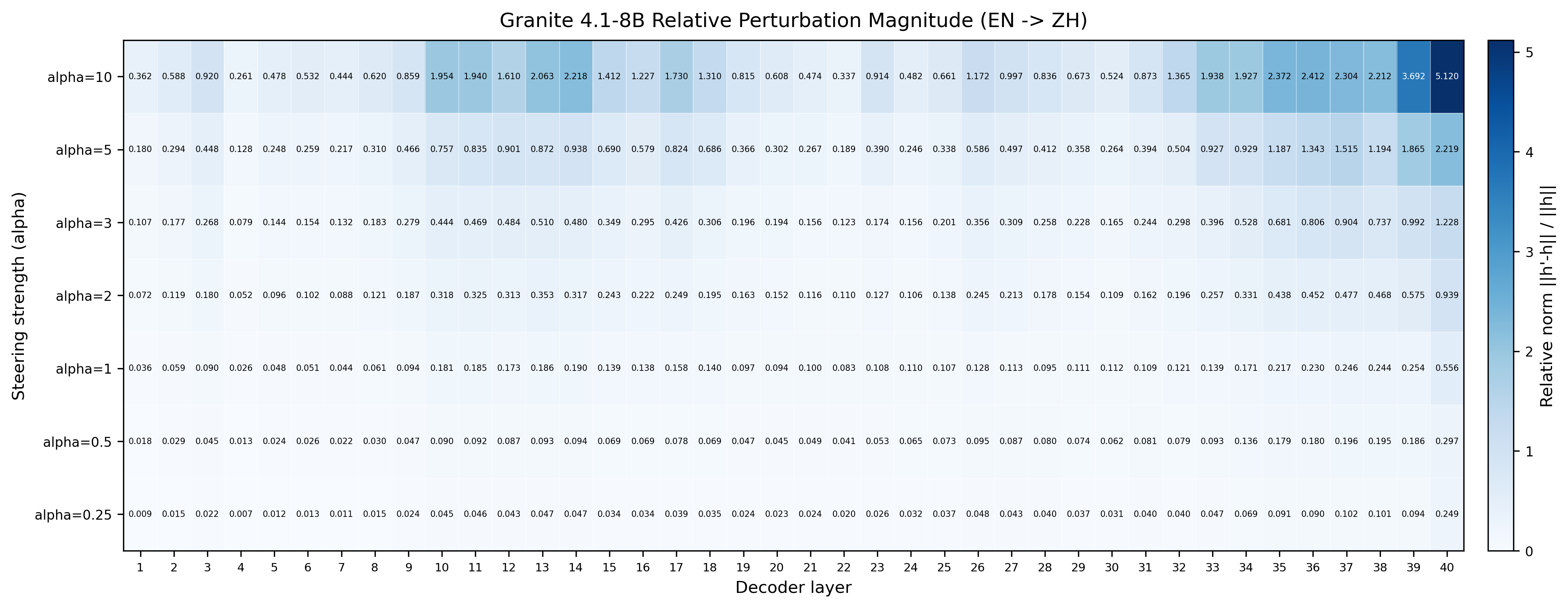}
    \caption{\textbf{Granite-4.1-8B Perturbation Magnitude ($EN \rightarrow ZH$).}
    Relative change in the hidden-state norm across decoder layers and steering strengths.
    Moderate steering produces relatively small changes across much of the model, while stronger interventions can become much larger, particularly in later layers.}
    \label{fig:perturbation_granite}
\end{figure}

Together, these examples show that the same value of $\alpha$ can correspond
to very different intervention magnitudes across layers and models. This
motivates using the normalized perturbation magnitude $\rho$ when comparing
steering methods such as PCA and CAA.

\clearpage

\section{Chinese--Spanish Steering}

\begin{figure}[htbp]
    \centering
    \includegraphics[width=0.8\linewidth]{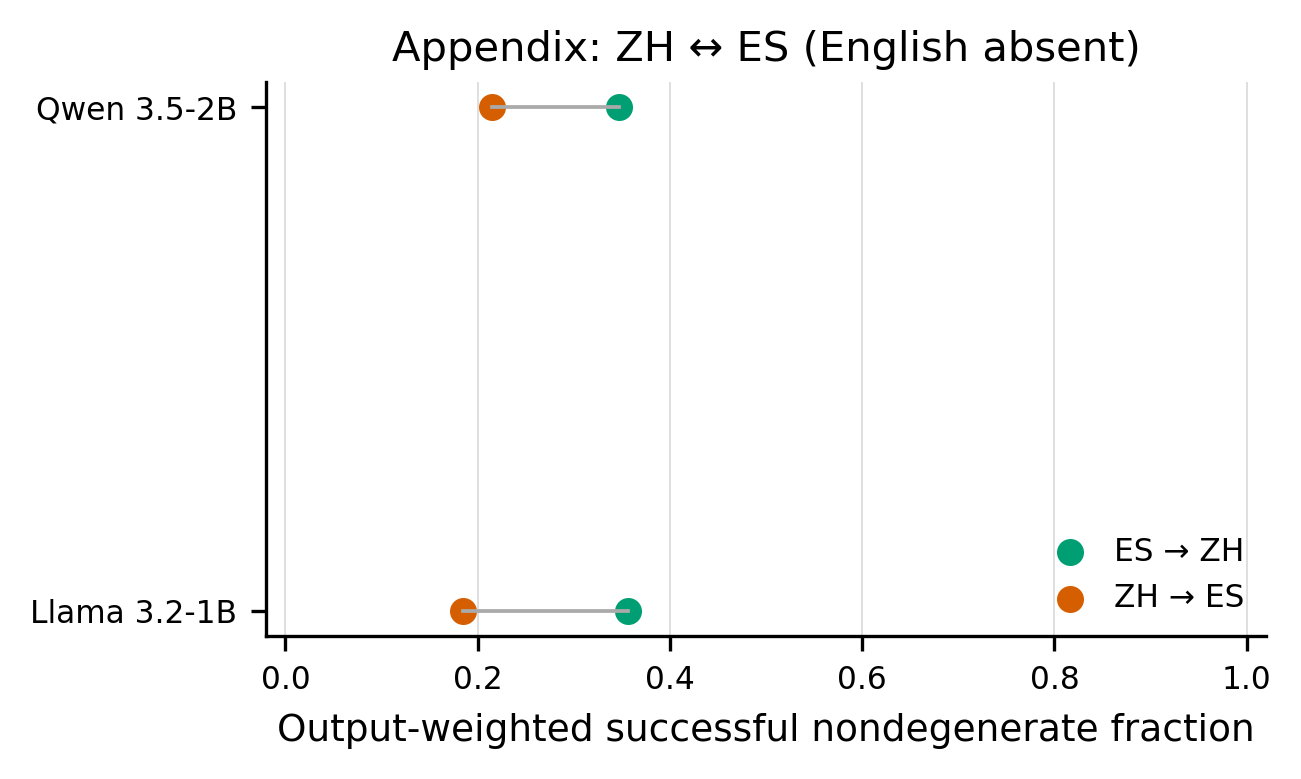}
\caption{\textbf{Chinese--Spanish directional comparison.}
Successful steering fraction for both Chinese--Spanish directions in
Qwen 3.5-2B and Llama-3.2-1B. Steering works in both directions, but
neither model shows the same consistent directional preference seen in
the English-centered experiments.}
    \label{fig:zh_es_directional}
\end{figure}

\begin{figure}[htbp]
    \centering
    \includegraphics[width=\linewidth]{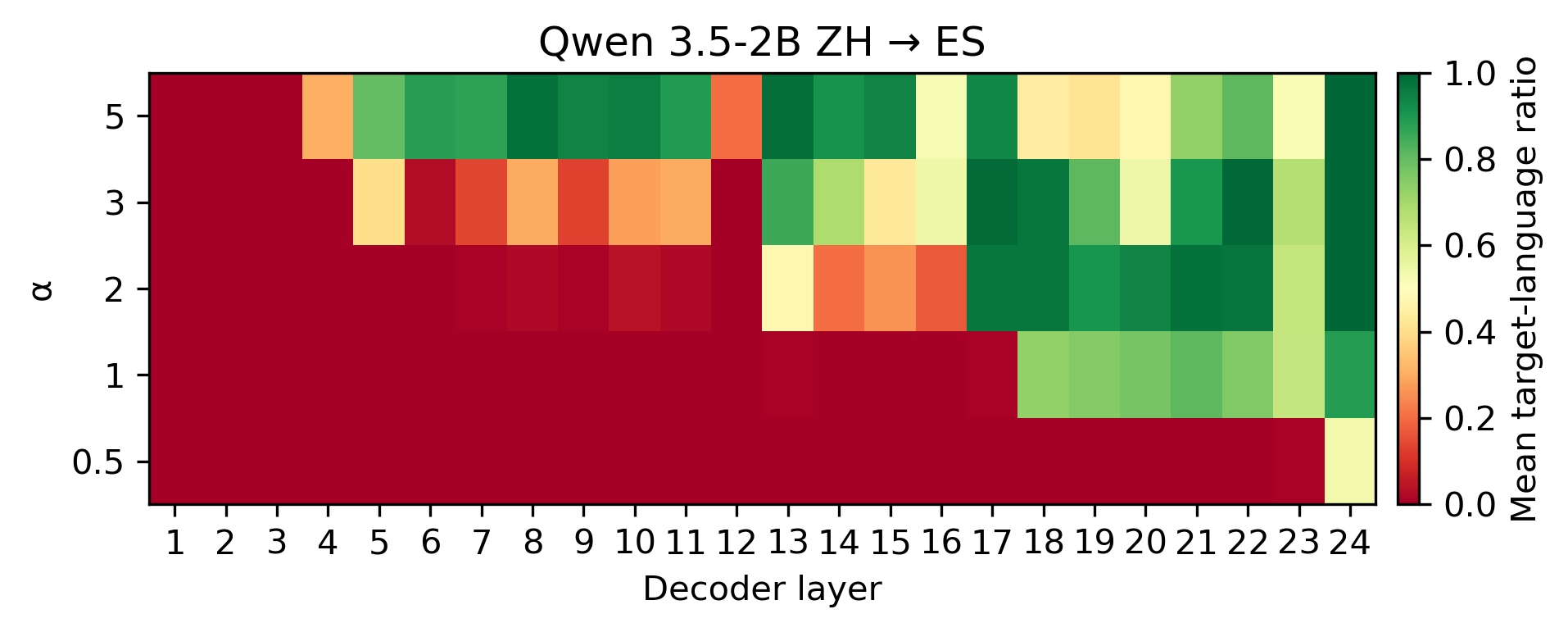}
    \caption{\textbf{Qwen 3.5-2B Chinese-to-Spanish Steering.}
    Target-language ratio across decoder layers and steering strengths.
    Steering becomes effective across several middle and later layers, showing that the language-axis intervention remains effective even when English is not part of the language pair.}
    \label{fig:heatmap_zh2es_qwen}
\end{figure}

\clearpage

\section{Targeted Qwen 3.5-27B Results}

\begin{figure}[htbp]
    \centering
    \includegraphics[width=0.75\linewidth]{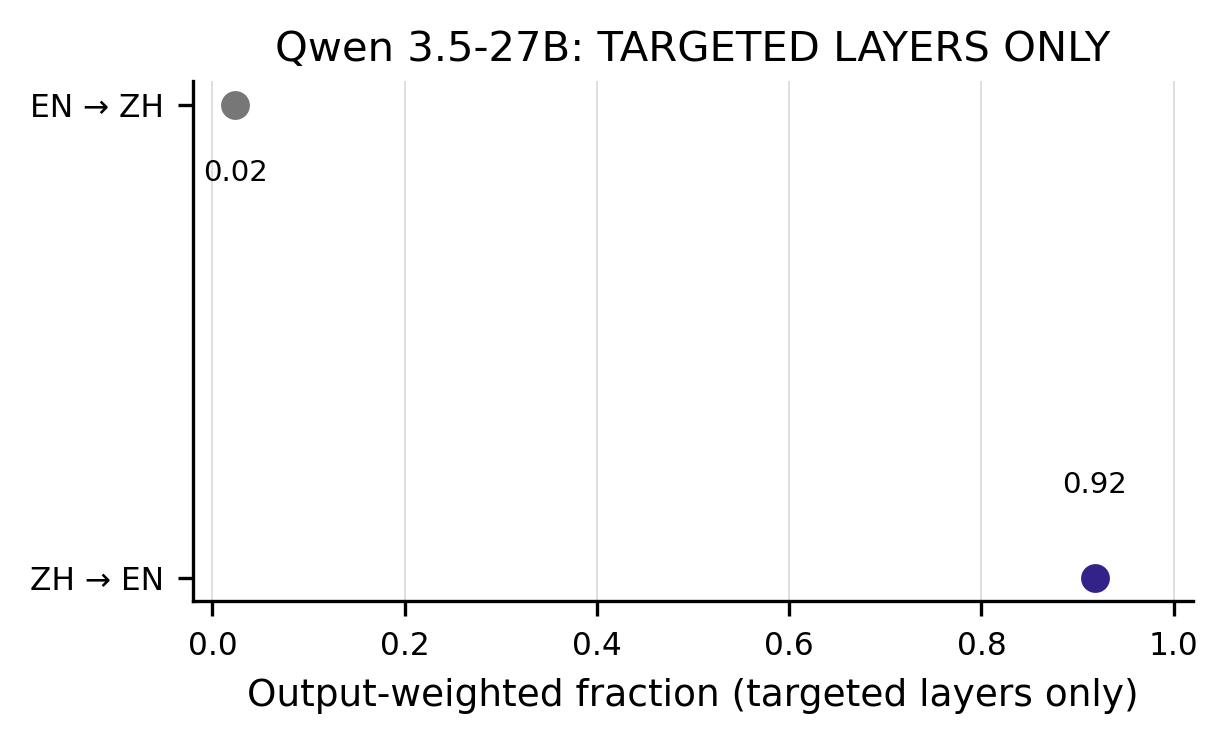}
\caption{\textbf{Targeted Qwen 3.5-27B steering.}
Successful steering fraction across the selected layers tested in each
English--Chinese direction. Steering toward English remains strong in
the targeted experiment, while steering toward Chinese is substantially
weaker. Because only selected layers were tested, we treat this as
additional evidence at larger scale rather than a full-depth comparison.}
    \label{fig:qwen27_targeted}
\end{figure}

\clearpage
\section{Prompting vs. steering}

\begin{figure}[htbp]
    \centering
    \includegraphics[width=\linewidth]{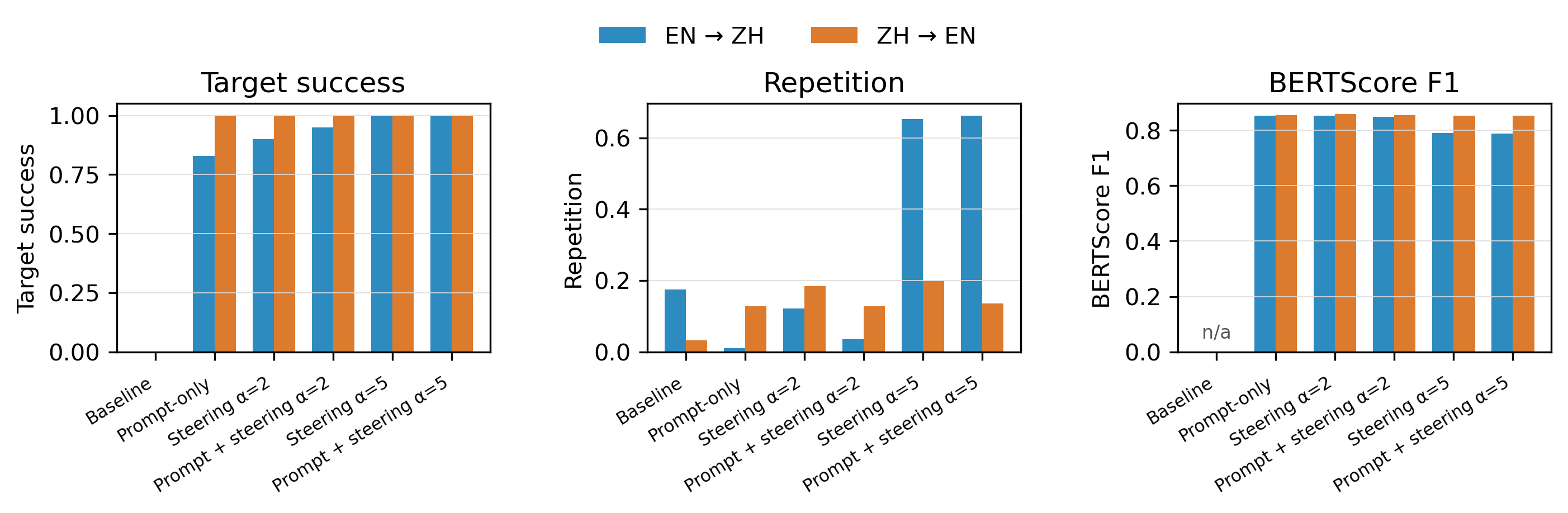}
    \caption{\textbf{Prompting versus activation steering in Qwen 3.5-2B.}
    Comparison of explicit prompting, activation steering, and their combination for $EN \rightarrow ZH$ and $ZH \rightarrow EN$.
    The panels report target-language success, repetition, and BERTScore F1.
    Prompting is already effective, while steering can further improve language control in the harder $EN \rightarrow ZH$ direction, especially when combined with prompting.}
    \label{fig:prompting_vs_steering}
\end{figure}

\clearpage
\section{Semantic Evaluation Details}

\begin{figure}[htbp]
    \centering
    \includegraphics[width=\linewidth]{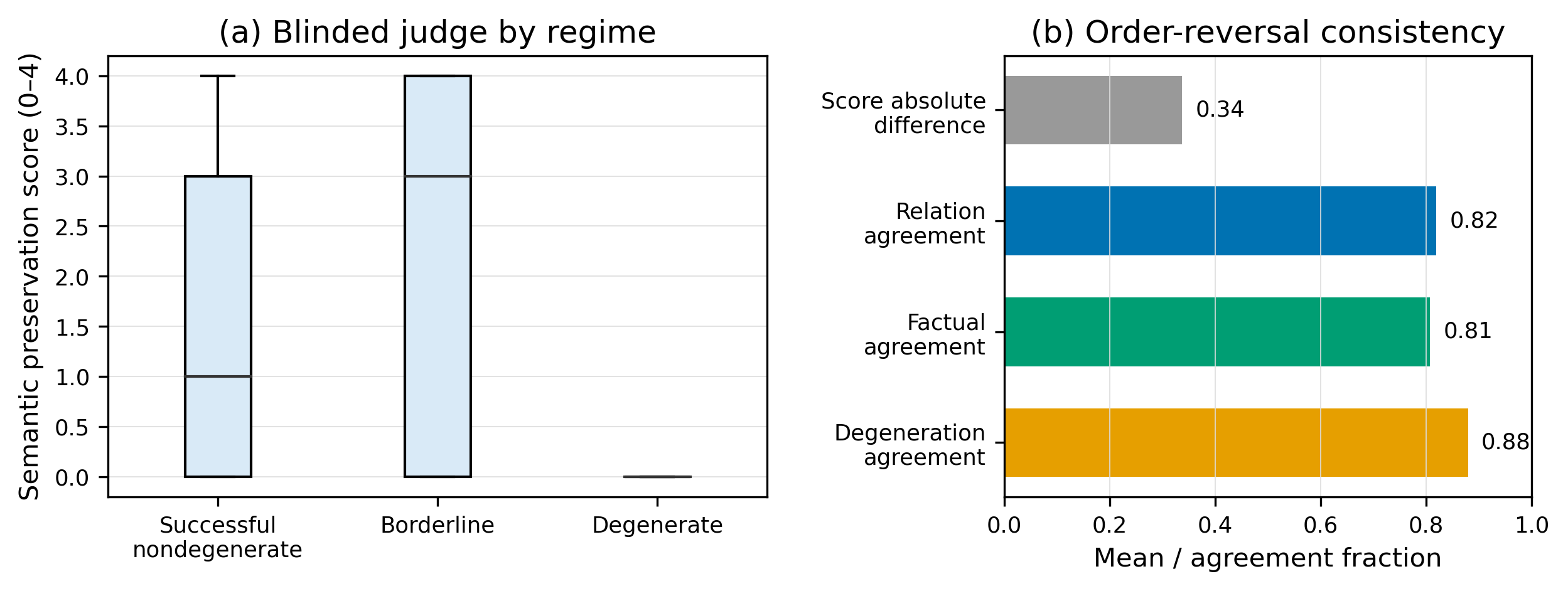}
    \caption{\textbf{Blinded semantic-evaluation diagnostics.}
    (a) Semantic-preservation scores across intervention regimes.
    (b) Consistency when the order of the two compared responses is reversed.
    The judge shows high agreement on semantic relation, factual consistency,
    and degeneration under order reversal.}
    \label{fig:llm_judge_diagnostics}
\end{figure}

\clearpage

\section{PCA and CAA Comparison}

\begin{figure}[htbp]
    \centering
    \includegraphics[width=0.85\linewidth]{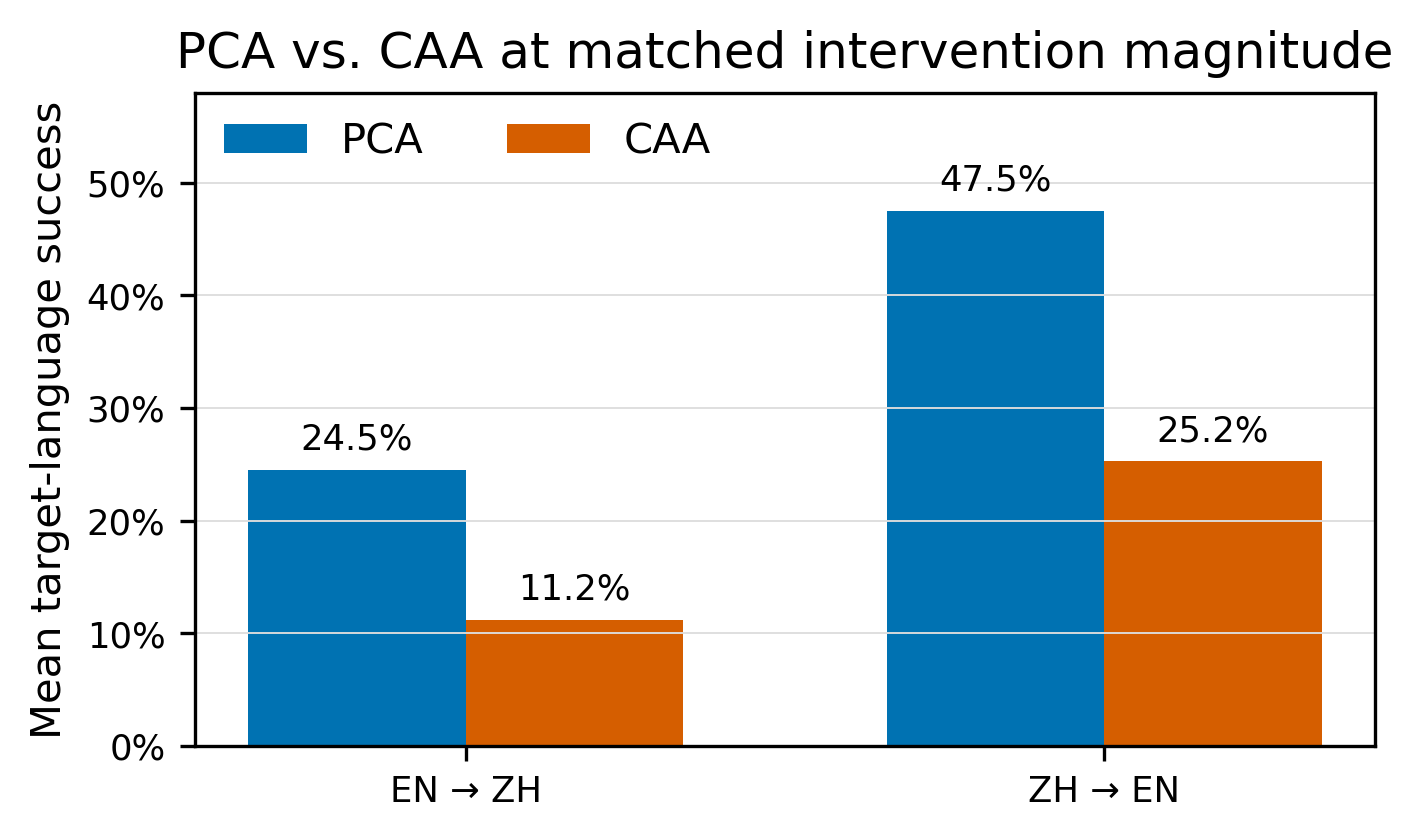}
    \caption{\textbf{PCA versus CAA at matched intervention magnitude.}
    Mean target-language success for Qwen 3.5-2B after matching PCA and CAA by their observed perturbation magnitude.
    PCA achieves higher target-language success in both $EN \rightarrow ZH$ and $ZH \rightarrow EN$ under this matched comparison.}
    \label{fig:pca_caa_matched}
\end{figure}

\end{document}

%% file: math_commands.tex
\usepackage{amsmath,amsfonts,bm}

\def\eqref#1{equation~\ref{#1}}

\def\1{\bm{1}}

\DeclareMathAlphabet{\mathsfit}{\encodingdefault}{\sfdefault}{m}{sl}
\SetMathAlphabet{\mathsfit}{bold}{\encodingdefault}{\sfdefault}{bx}{n}

